\documentclass[preprint,authoryear,a4paper]{elsarticle}

\usepackage{verbatim}

\usepackage{geometry}

\usepackage{color,soul}

\usepackage[hidelinks]{hyperref}
\hypersetup{
  colorlinks   = true, 
  urlcolor     = blue,
  linkcolor    = blue, 
  citecolor    = blue 
}

\usepackage{graphicx}
\graphicspath{{./images/}}
\DeclareGraphicsExtensions{.pdf,.png,.jpg}
\usepackage{placeins}
\usepackage{caption}
\usepackage{subcaption}

\usepackage{booktabs}
\usepackage[table,xcdraw]{xcolor}
\usepackage{graphicx}
\usepackage{adjustbox}
\usepackage{multirow}

\usepackage{amssymb}
\usepackage{gensymb}
\usepackage[mathscr]{euscript}
\usepackage{bm}
\usepackage{mathtools}
\usepackage{nccmath}
\usepackage{amsmath}
\usepackage{braket}

\usepackage{algorithm}
\usepackage{algpseudocode}

\usepackage{lineno, hyperref}

\usepackage{lipsum}

\journal{ISPRS Journal of Photogrammetry and Remote Sensing}

\begin{document}

\begin{frontmatter}

    \title{Semi-Supervised Adversarial Recognition of Refined Window Structures for Inverse Procedural Fa\c{c}ade Modeling}

    \author[swjtu]{Han Hu}
    \author[swjtu]{Xinrong Liang}
    \author[swjtu]{Yulin Ding\corref{cor1}}
    \author[swjtu]{Xuekun Yuan}
    \author[swjtu]{Qisen Shang}
    \author[swjtu]{Bo Xu}
    \author[swjtu]{Xuming Ge}
    \author[swjtu]{Min Chen}
    \author[bnu]{Ruofei Zhong}
    \author[swjtu]{Qing Zhu}
    \cortext[cor1]{Corresponding Author: dingyulin@swjtu.edu.cn}

    \address[swjtu]{Faculty of Geosciences and Environmental Engineering, Southwest Jiaotong University, Chengdu, China}
    \address[bnu]{Beijing Advanced Innovation Center for Imaging Technology, Capital Normal University, Beijing, China}
    \begin{abstract}     
		Deep learning methods are notoriously data-hungry and requires many labeled samples.
		Unfortunately, the interactive sample labeling efforts have dramatically hindered the application of deep learning methods, especially for the 3D modeling tasks, which require heterogeneous samples.
		This paper proposes a semi-supervised adversarial recognition strategy embedded in the inverse procedural modeling to alleviate the work of data annotation for the learned 3D modeling of fa\c{c}ades.
		Beginning with textured LOD-3 (Level-of-Details) models, we use the convolutional neural networks to recognize the types and estimate the parameters of windows from image patches.
		The window types and parameters are then assembled into the procedural grammar.
		A simple procedural engine is built inside an off-the-shelf 3D modeling software, producing fine-grained window geometries.
		To obtain a useful model from few labeled samples, we leverage the generative adversarial network to train the feature extractor in a semi-supervised manner.
		The adversarial training strategy exploits the unlabeled data to make the training phase more stable.
		Experiments using publicly available fa\c{c}ade image datasets reveal that the proposed methods can obtain about a 10\% improvement in classification accuracy and a 50\% improvement in parameter estimation under the same network structure.
		In addition, performance gains are more pronounced when testing against unseen data featuring different fa\c{c}ade styles.
		
    \end{abstract}

    \begin{keyword}
        Oblique Photogrammetry \sep 3D Building Model \sep Generative Adversarial Network \sep Semi-Supervised Learning \sep LOD3 \sep Inverse Procedural Modeling
    \end{keyword}
\end{frontmatter}

\section{Introduction}
\label{s:intro}

Buildings form the basic structures of the urban environment, and fa\c{c}ades are the most prominent features of the street view.
With the advent of aerial oblique images and ground mobile mapping systems, city-scale photo-realistic 3D models with fa\c{c}ade visibility are readily available \citep{verdie2015loda,han2021urban}.
Detailed analysis and parsing of the fa\c{c}ades \citep{fan2021layout,hu2020fast}, which enrich the 3D models with semantic information conformal to the LOD-3 (Level-of-Details) protocol in CityGML \citep{groger2012citygml}, have recently raised extensive attentions in the community.
Although the LOD-3 models already contain computational information for various applications, the window taxonomy and refined structures are often ignored \citep{zhu2020large}.
The detailed geometrical information associated with texture materials is crucial for realistic representations of the 3D window models.
In addition, the semantic information of refined window taxonomy also enables advanced applications in urban management.
This paper focuses on the refined window modeling based on the LOD-3 models to improve the realistic visualization and to support advanced applications, such as virtual reality and detection of illegal fa\c{c}ade construction.

Point clouds \citep{zolanvari2018threedimensional,nan2010smartboxes}, photogrammetric mesh models \citep{verdie2015loda} and images \citep{susannebecker2007combined} are probably the most widely used data sources for reconstructing the refined structures of windows.
For the inner parts of windows, point clouds are represented as void regions \citep{fan2021layout,dehbi2017statistical} and photogrammetric mesh models are defects-laden \citep{verdie2015loda}; therefore, images are probably the best option.
Two common strategies are available \citep{wang20133d} for geometrical modeling of windows, e.g., the data-driven and model-driven approaches.
Because the types of windows in a particular region are generally limited, we adopt the model-driven approaches, similar to recognizing building roof types and estimating roof parameters \citep{huang2013generative,xu2021efficient}.

In this regard, we pose the problem of detailed window reconstruction as image classification \citep{simonyan2015very} for the window types and regression \citep{ren2015faster} for the parameters.
Despite the recent progress on the learned 3D modeling \citep{kelly2018frankengan,nishida2018procedural,zhang2020convmpn}, the notorious data-hungry issue for deep learning still impedes the adoption of such approaches in practice.

\begin{figure}[t]
	\centering
	\subcaptionbox[a]{ImageNet-like classes vs. window types with 100 labeled samples}[0.48\linewidth]{\includegraphics[width=\linewidth]{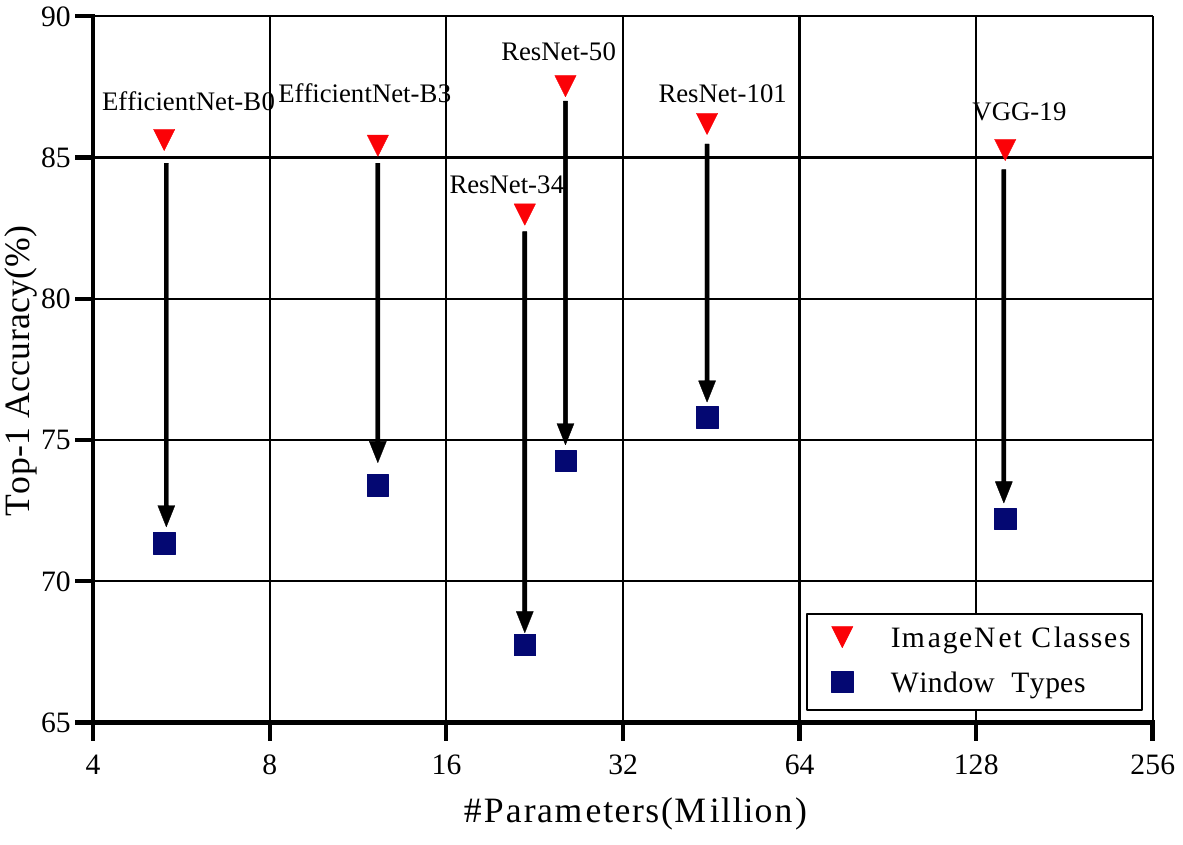}}
	\subcaptionbox[b]{Different training samples}[0.48\linewidth]{\includegraphics[width=\linewidth]{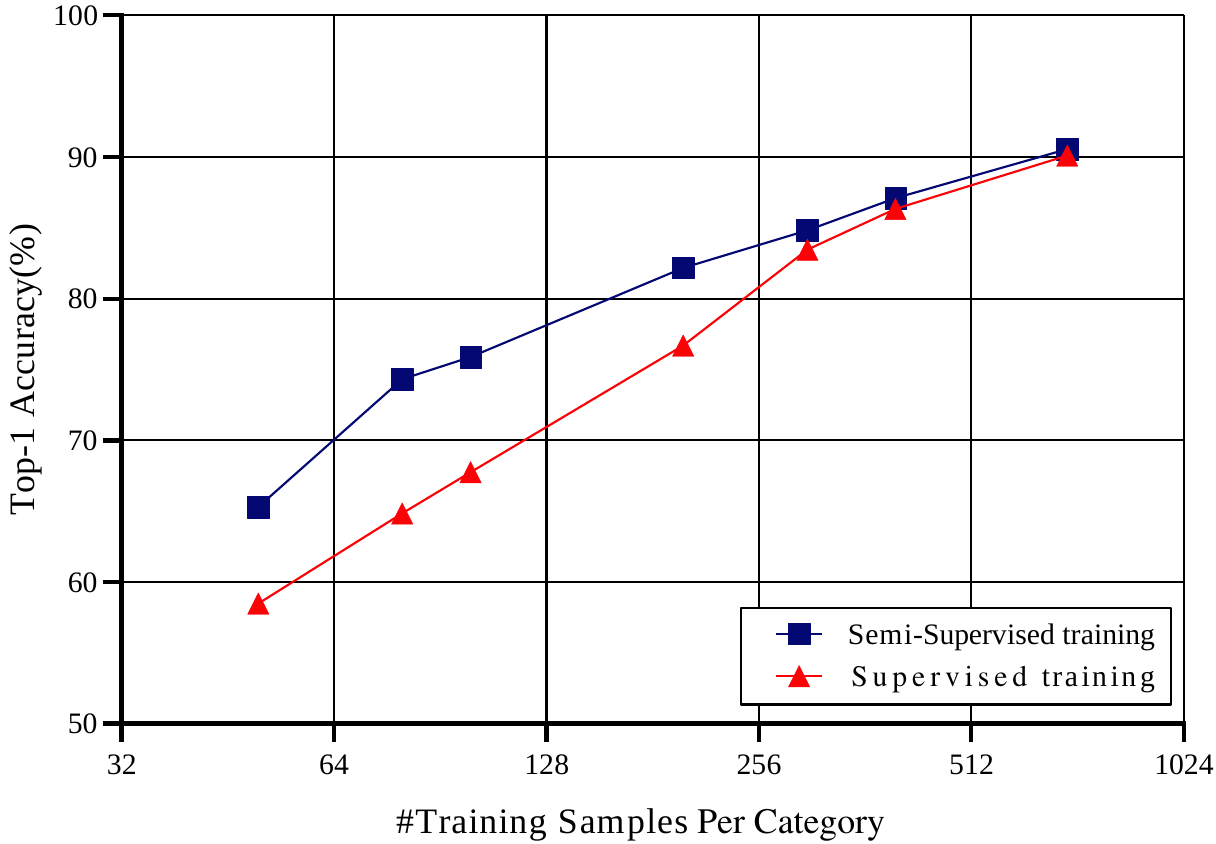}}
	\caption{Problems with recognizing fine-grained window types using a small number of labeled data and pre-trained ResNet. (a) Image classification accuracy is significantly higher for nine ImageNet classes than nine window types during fine-tuning using only 100 labeled samples per category. (b) With decreasing number of labeled data per category (from 900 to 50), the accuracies for recognizing window types decreased steeply with only supervision from the labeled samples, which is alleviated by semi-supervised training with additional $1.6\times10^5$ unlabeled window patches.}
	\label{fig:train}
\end{figure}

\textit{1) Limited transferring capability between different domains with sparse labeled samples.}
For image classification, i.e., recognizing the input image to different categories, transfer learning \citep{deng2009imagenet} is a well-known strategy to achieve faster convergences and more stable training results.
However, when the amount of labeled training data is small, fine-tuning to ImageNet-like classes shows significantly better results than fine-grained window types (Figure \ref{fig:train}a), i.e., recognition between windows with different styles.
The above issue is probably because of the domain gap phenomenon \citep{xie2020pointcontrast}: the detailed structures of different window types are embedded into the same distribution by the pre-trainied models.

\textit{2) Infeasible results from only sparse training samples.}
In practice, we cannot predetermine all possible configurations of window types, and collecting project-specific samples is inevitable.
Because labeling training samples is time-consuming and sometimes even impossible, a smaller number of samples is preferred.
Unfortunately, by decreasing the labeled data, the training procedural is unstable \citep{chen2020simple} and prone to over-fitting, with a steep performance loss (Figure \ref{fig:train}b).

This paper proposes a semi-supervised recognition strategy with only few labeled samples to solve the two issues above by additionally utilizing the abundant unlabeled samples.
We add a Generative Adversarial Network (GAN) \citep{goodfellow2014generative} branch parallel to the backbone network to improve training stability from few labeled samples.
Firstly, the GAN with an autoencoder structure is used to generate realistic window images, which enforces the feature extractor fitting the data inside the window domain; in this way, we can better transfer the learned weights from ImageNet to window domain.
Secondly, the generated window images also serve as pseudo training samples to improve the stability of the training procedure with few labeled data in a semi-supervised manner \citep{odena2016semisupervised}.
Specifically, we first design several parametric window structures procedurally created in a 3D modeling software \citep{trimble2021sketchup}.
Then, the proposed adversarial training strategy learns a convolutional neural network (CNN) to recognize different window types in the training stage.
We also estimate the corresponding parameters for each class through the same network structure.
Finally, the network models serve as a recommendation system to suggest different window types and parameters for each window or a set of grouped windows.

In summary, this paper contains two contributions concerning the data-hungry issue for learned 3D modeling of refined window structures: 1) an additional semi-supervised pre-training stage to improve the performance of transfer learning and 2) an adversarial learning strategy to improve the accuracy and stability with few labeled samples.
The rest of this paper is organized as below.
Section \ref{sec:related_work} provides a brief review of related works.
Section \ref{sec:approach} elaborates the details of the proposed method.
Experimental evaluations and analyses are presented in Section \ref{sec:experiments}.
Finally, the last section concludes the paper.

\section{Related works}
\label{sec:related_work}
In the following, we briefly review the most relevant topics in the literature, including fa\c{c}ade modeling and generative adversarial networks.

\textbf{Fa\c{c}ade modeling.}
Earlier works on fa\c{c}ade modeling generally used low-level features to estimate the fa\c{c}ade structures.
For instance, contours of void regions \citep{susannebecker2007combined,pu2009building} and line segments from the images \citep{pu2009building,tian2010knowledgebased} are common cues for the window details.
The planar point clouds segments enriched with geometric and radiometric features were also helpful for generating semantic information of fa\c{c}ade \citep{dehbi2017statistical,wang2018accurate,li2017hierarchical}.
Because of the semantic gap between low-level features and high-level representations, the above works relied heavily on the \textit{a priori} knowledge of the fa\c{c}ade and suffered severely from the noise in the data.

Due to the impressive feature learning capability \citep{bengio2013representation}, deep learning approaches are also widely adopted for fa\c{c}ade modeling.
Two typical strategies in image processing are adopted, e.g., object detection \citep{ren2015faster} and semantic segmentation \citep{long2015fully}.
For the former, the rectangular regions of the windows are detected using a typical detector \citep{ren2015faster} and used for fa\c{c}ade reconstruction \citep{hu2020fast,hensel2019facade,kong2020enhanced}; however, there is still ample space to improve the localization accuracy, e.g., the intersection of union (IoU).
For the latter, pixel-wise segmentation results can be learned end-to-end \citep{mathias2016atlas,gadde2018efficient}; fusing with point clouds \citep{gadde2018efficient} and multi-view voting \citep{ma2020multiview} can also improve the segmentation results.
In addition, the combination of the two strategies (termed as instance segmentation) \citep{he2017mask}, which detects the bounding boxes first and conducts pixel-wise segmentation inside each object, is also a popular approach \citep{liu2020deepfacade}.

The detection and segmentation approaches above generally cannot enforce the geometry regularities of the fa\c{c}ade objects, such as the alignment and equal-size constraints.
A straightforward approach for regularization explicitly offsets adjacent points or boundaries in a least-squares optimization \citep{arikan2013osnap,zhu2020interactive,xie2018hierarchical}.
Because least-squares optimization cannot model the logic operators, such as \emph{sameColumn} and \emph{sameWidth} \citep{dehbi2017statistical}, clustering \citep{fan2021layout}, learned Markov Logic Networks \citep{dehbi2017statistical}, or binary integer programming \citep{hu2020fast,hensel2019facade,monszpart2015rapter} are exploited to snap the feature points on the fa\c{c}ade.

Most of the existing work on fa\c{c}ade modeling does not recover the refined structures. 
The following works on detailed modeling of buildings using a learned approach are the most relevant to our work.
\cite{nishida2018procedural} annotated $10^5$ images and recovered different window grammars; however, the parameters in the procedural grammars were still kept fixed due to the infeasible amount of labeling work.
\cite{zeng2018neural} categorized the building foundation, roof, and garage using the deep CNNs from normal and height maps rendered using LiDAR (Light Detection and Ranging) point clouds.
Procedurally generated models using CityEngine \citep{esri2021arcgis} were also used to synthesize the normal and height maps as the training samples.
Unfortunately, the generation of photo-realistic window patches with fine-grained control over the types and parameters is not as straight-forward as the height and normal maps in the work by \cite{zeng2018neural}.

\textbf{Generative Adversarial Networks.}
GAN structures consist of two components: a generator and a discriminator \citep{goodfellow2014generative}.
The generator creates \textit{fake} samples from random signals, which try to follow the same distribution as the training data.
The discriminator examines the generated samples and predicts whether it comes from actual data or random signals.
The parameters in the two components are learned through a minimax game.
The game looks for the Nash equilibrium rather than a (global) minima for an optimization problem \citep{salimans2016improved}.
After the seminar work \citep{goodfellow2014generative}, DCGAN (Deep Convolutional GAN) \citep{radford2016unsupervised} proposed a generator to create high-quality images, using the deconvolution operator \citep{springenberg2015striving,long2015fully}.
Concerning the discriminator part, there are also many improvements, notably the CutMix augmentation \citep{yun2019cutmix}, patch-wise \citep{isola2017image} and even pixel-wise \citep{schonfeld2020unet} penalization.

GANs are capable of capturing the data distribution and emulating the training dataset.
They are naturally helpful for many tasks that intrinsically require the realistic generation of samples or images \citep{radford2016unsupervised} in a specific domain.
In single image super-resolution, generative processing can inject more information into the synthesized high-resolution image \citep{ledig2017photorealistic}.
GANs can also assist the image completion problem \citep{zhao2021large} for similar reasons, such as artwork creation \citep{iizuka2017globally} and thin cloud removal \citep{li2020thin}.
Another exciting direction is the image-to-image translation \citep{isola2017image}, which maps an observed image to a different domain through conditional GANs \citep{mirza2014conditional}; the generator and discriminator are additionally conditioned on an observed image.

It is universally acknowledged that deep learning algorithms are data-hungry.
Weakly supervised \citep{salimans2016improved} or unsupervised \citep{radford2016unsupervised} approaches have become popular to conquer the fundamental generalization problem with few labels.
GANs are inherently a powerful tool for this task \citep{goodfellow2014generative}.
Intuitively, the weights in the discriminator can be learned from the task of classifying the real-vs-fake images \citep{radford2016unsupervised,donahue2019large}.
Because GANs are better trained with samples in a particular form \citep{goodfellow2017nips,salimans2016improved}, semi-supervised GAN (SGAN) was proposed \citep{salimans2016improved,odena2016semisupervised}. 
More specifically, the discriminator exploited a pseudo-class representing the fake samples in addition to the original classes and enabled training with unlabeled data in the target domain \citep{salimans2016improved,odena2016semisupervised}.

This paper embraces GANs for semi-supervised learning of refined window structures.
A significant amount of unlabeled window images enforces the model fitting inside the distribution of windows.
This feature benefits the task of recognizing detailed window structures and estimating the procedural parameters by fine-tuning a pre-trained model.

\section{Methods}
\label{sec:approach}
\subsection{Overview and problem setup}
\label{ssec:overview}

\subsubsection{Overview of the approach}

\begin{figure}[t]
	\centering
	\includegraphics[width=\linewidth]{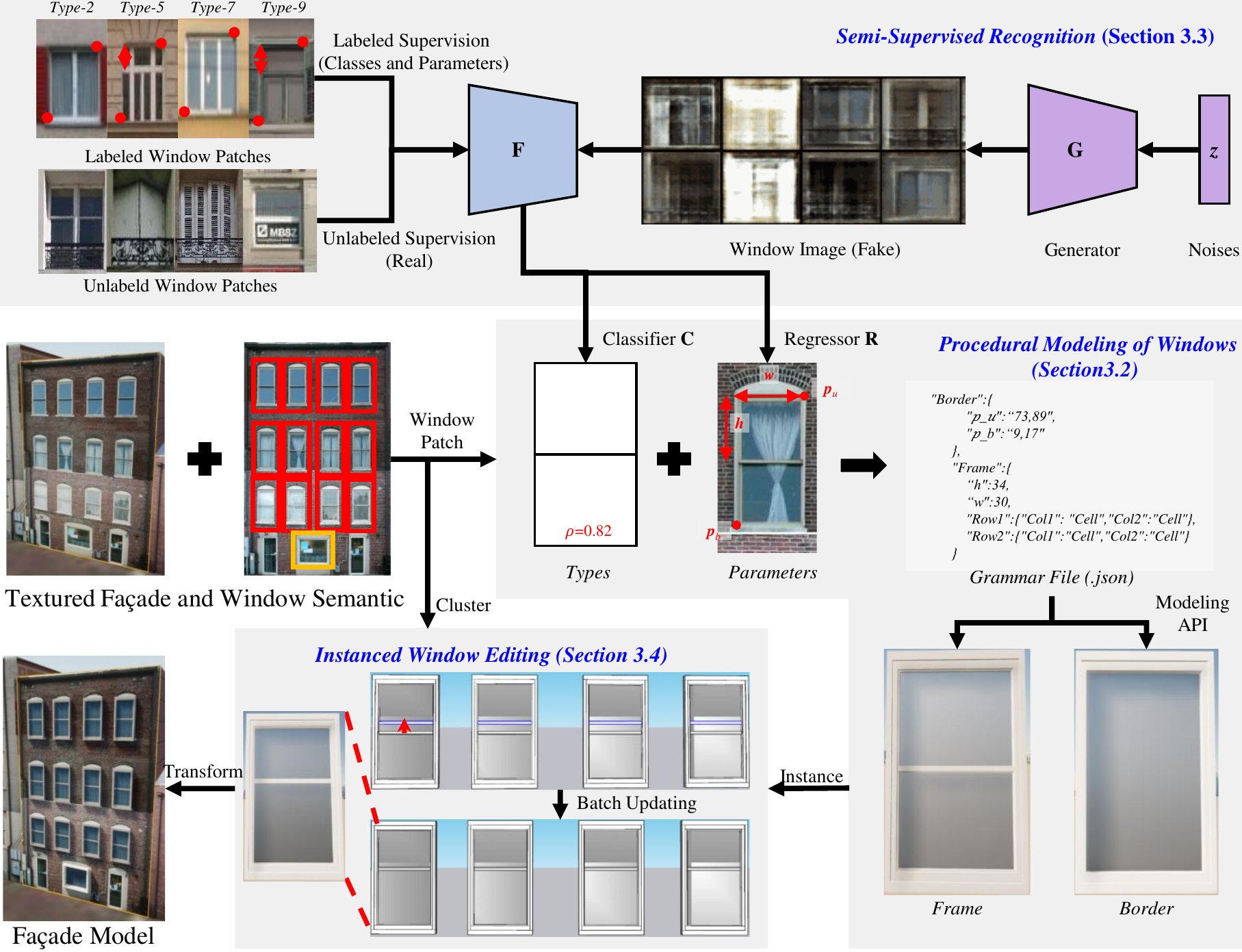}
	\caption{Overview of the proposed fa\c{c}ade modeling system.}
	\label{fig:overview}
\end{figure}

We provide an overview of our fa\c{c}ade modeling pipeline in Figure \ref{fig:overview}.
The texture of each wall should be continuous in the atlas space rather than discrete as for photogrammetric mesh models \citep{zhu2021structureaware}.
In contrast to FrankenGAN \citep{kelly2018frankengan}, which generated high-resolution window texture and detailed window geometries guided by an input style, our objective is to reconstruct realistic 3D window geometries well aligned with the appearances of existing high-resolution images.
Therefore, this study is not formulated as an entire generated problem.
In addition, less labeled data is required.
We assume the details of windows are visible and apparent in the texture image.
Different windows can be grouped into clusters sharing the same type and parameter, and the instanced modeling API (Application Programming Interface) supports batch editing for models in the same set (Section \ref{ssec:instancing}).

We define a simple shape grammar represented in a tree structure, similar to the CGA (Computer Generated Architecture) approach \citep{muller2006procedural}.
Implementing a procedural modeling engine \citep{muller2006procedural,esri2021arcgis} for a complete set of shape grammar is non-trivial.
Since we only target the windows, the shapes can be conveniently generated procedurally using the modeling API provided by a 3D modeling solution \citep{trimble2021sketchup}, as elaborated in Section \ref{ssec:procedural}.
To facilitate the inverse generation of shape grammars from the image, we abstract a certain number of window types, learn the styles from the cropped image, infer the parameter associated with the window type (Section \ref{ssec:classification}), and finally produce the parameterized shape grammar.
The 3D window instances are generated automatically and assembled with the corresponding transformation information.

\subsubsection{Problem Setup}
More formally, the input is a polygonal surface mesh consisting of $F$ faces, e.g., $\mathcal{M}=\{\mathcal{F}_i\}$.
Each face is associated with an individual texture image $\mathcal{I}$ with the axis-aligned bounding boxes  $\mathcal{B}$ indicating the windows.
For each image patch cropped from the bounding box $\mathcal{B}_i$, we generate a tree grammar $\mathcal{G}_i$ that consists of hierarchical nodes of parameterized elements, including \textit{frame}, \textit{row}, \textit{column}, and \textit{cell}.
To ease the learning process, we abstract the shape grammar into two consecutive steps for the types and geometry parameters, and a unified shape grammar suffices for different window types.
We interactively assign similar windows into the same cluster of geometric instances $\mathcal{C}=\{\mathcal{G}_i\}$; the parameters of the shape grammar for the set of geometry instances $\mathcal{C}$ are averaged.
The multiple instances in the same cluster share the underlying geometrical objects.
Multiple window mesh models can be instantiated with the associated transformation matrices at different locations.
In addition, the instanced modeling API \citep{trimble2021sketchup} supports batch editing for models in the same cluster.
A lightweight procedural modeling engine turns the shape grammar into the geometry instance.
By this mechanism, changes on any instance will immediately update all the window models.
Given the correct transformation matrices, all the window models will be placed on the corresponding wall faces $\mathcal{F}$, forming the final output.

\subsection{Procedural Modeling of Windows}
\label{ssec:procedural}

Unlike the model-driven approaches that use a fixed set of parametric primitives \citep{henn2013model}, we exploit the shape grammar \citep{muller2006procedural} to define the structures of windows and generate the 3D models procedurally.
Although the wall layout was expressed using the complex operators in the split grammars or CGA rules, the detailed geometry of windows was still ignored in previous works \citep{muller2006procedural,nishida2016interactive}, as a simple insertion operation was adopted with an external window model.
Because the parameters for the detailed windows were not adjustable, a significant number of fixed window types were required, reducing the recognition accuracy.
This study provides a more flexible pipeline, allowing interactions on procedurally generated windows.

\begin{figure}[h]
	\centering
	\subcaptionbox[a]{Snippets of the grammar}[\linewidth]{\includegraphics[width=0.8\linewidth]{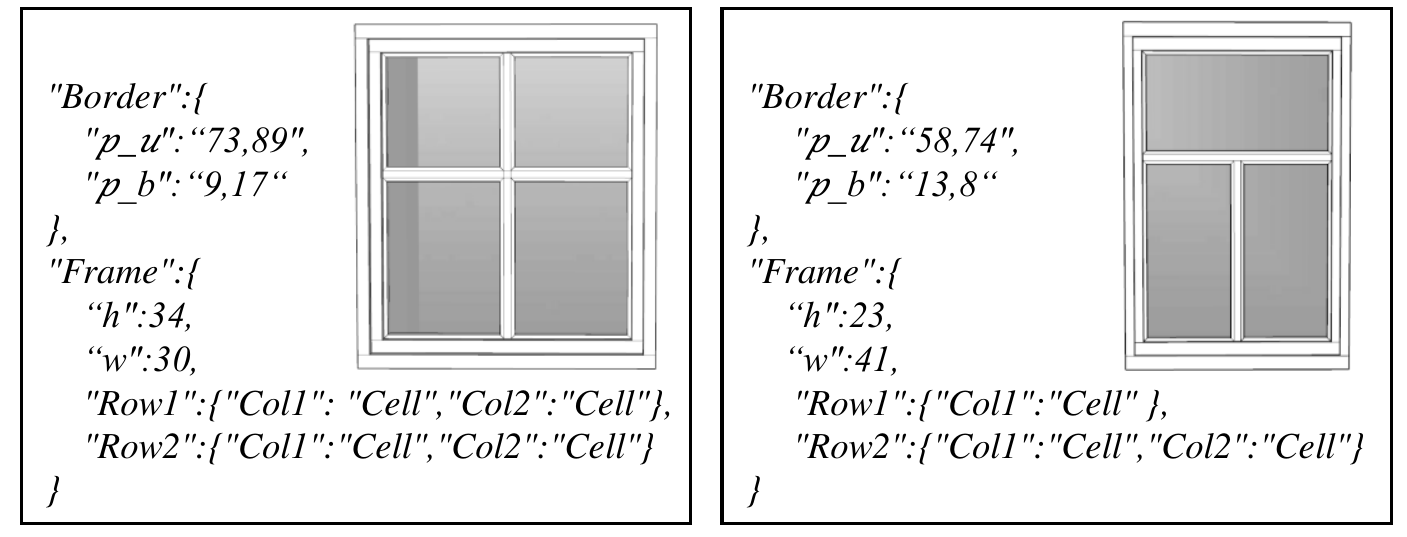}}
	\subcaptionbox[b]{Generation of the window instance}[\linewidth]{\includegraphics[width=\linewidth]{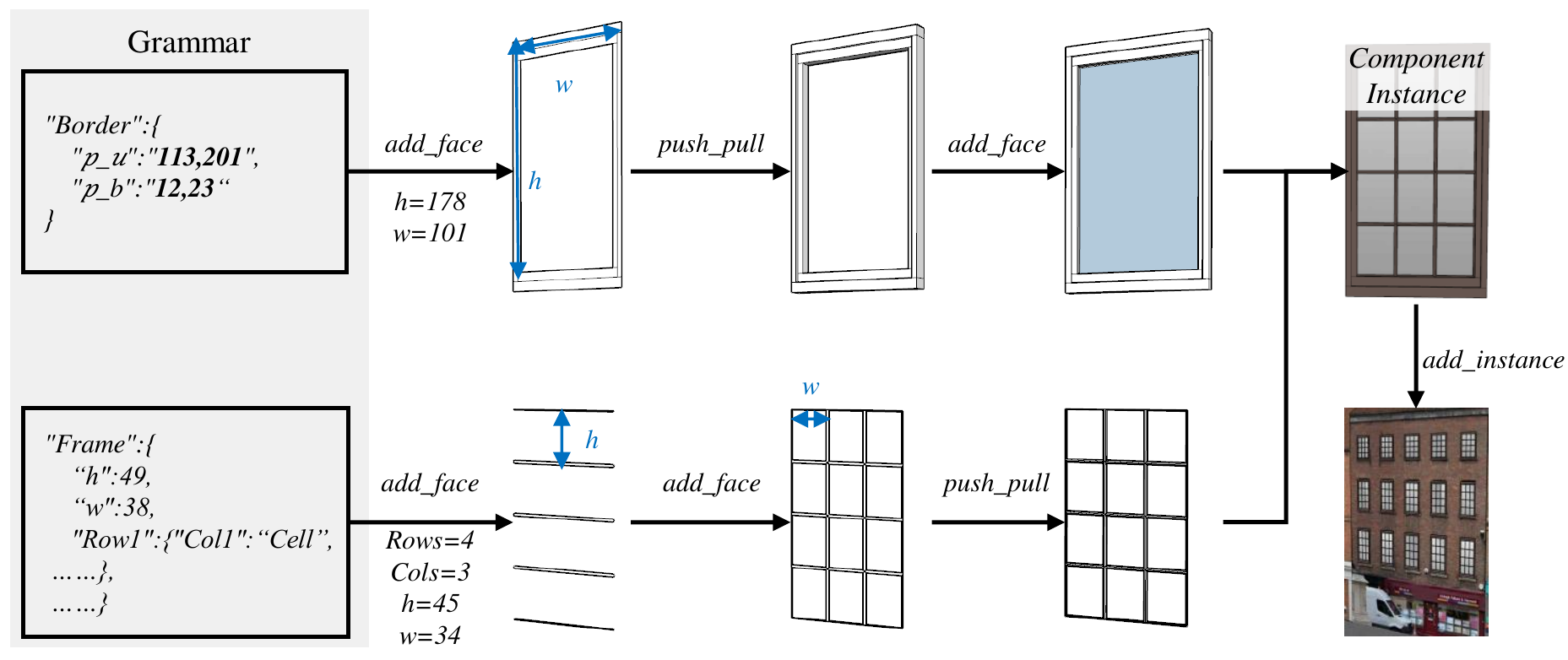}}
	\caption{Procedural modeling using the JSON-based grammar data. (a) Two examples of JSON-based snippets and the corresponding window models and (b) generation of the instances using the modeling API \citep{trimble2021sketchup}.}
	\label{fig:grammar}
\end{figure}

Our system implements a simple grammar represented as a JSON object inspired by previous work \citep{nishida2016interactive}.
All the window types can be represented using the same JSON-based grammar.
The grammar consists of the following elements.
(1) \textit{Border and frame}, the border is represented by a rectangular ring, and the frame is the inside part.
(2) \textit{Row and column}, the frame can be subdivided into intermediate rows and columns; row and column elements can be nested only once.
(3) \textit{Cell}, each border and frame must be terminated by a cell element, which finally produces a rectangular bar in the model.
Two examples of the shape grammar and the corresponding models are presented in Figure \ref{fig:grammar}a, as well as the window components in Figure \ref{fig:grammar}b.

\begin{figure}[htb]
	\centering
	\includegraphics[width=\linewidth]{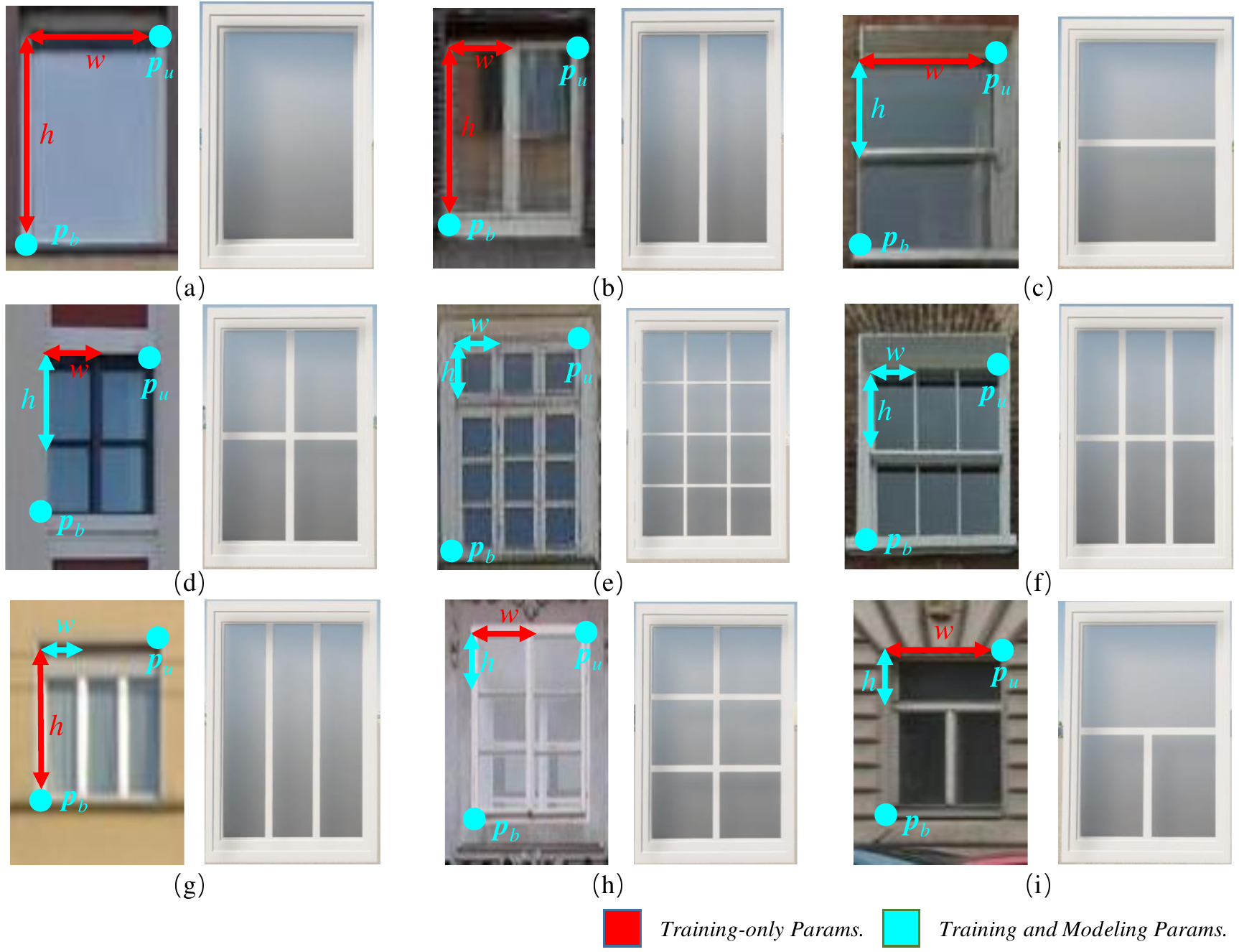}
	\caption{Window types and parameters used in the training of recognition and procedural modeling. Subfigures (a) through (i) illustrate different window types. The parameters in red are only used for training and the cyan parameters are also used in the modeling process.}
	\label{fig:types}
\end{figure}

The parameters include the border \textit{size} $\bm{s}$ (deducted from the two corner points as shown in Figure \ref{fig:types}), the row \textit{height} $h$ and the column \textit{width} $w$.
The two corner points $\bm{p}_u$ and $\bm{p}_b$ determine the window position relative to the image patch.
The image patch of the window is scaled anisotropically to a squared patch with 64 pixels for each edge.
We categorize the windows using the number of rows or columns, e.g., one, two, and multiple ($\ge$ three).
In this regard, nine window types are specified in this paper.
Figure \ref{fig:types} demonstrates window primitives considered in our system, including image examples, grammar parameters, and the corresponding learnable parameters.
To achieve plausible and regularized results, we assume all the columns have the same width, and all the cases for multiple rows have the same height.
Only windows with two rows can have varying heights.
Therefore, not all parameters in the training process are adopted for the 3D modeling.
In fact, for windows with multiple rows or columns, we learn the height and width of the cell and then calculate the number of rows or columns, respectively.
Because the assumption always produces regularized results, the algorithm can degrade gracefully even when the assumption is violated.

In summary, determining a shape grammar $\mathcal{G}_i$ requires a hierarchy of rows and columns for the frame and the associated parameters.
After generating a shape grammar $\mathcal{G}_i$ for a cluster of windows, we can create a geometry instance $\mathcal{C}_i$ using the modeling API.
More specifically, we first infer the location and size of each cell.
Then a rectangular ring is generated for each cell using the \textit{offset} and \textit{extrude} operations \citep{trimble2021sketchup}, as shown in Figure \ref{fig:grammar}.
Multiple windows belonging to the same cluster are cloned to the corresponding locations.
After the automatic creation of the geometry instance of a cluster $\mathcal{C}_i$, we can edit the instance $\mathcal{C}_i$ interactively to better align with the overlaying image, including the geometries and materials of the frames and glasses.

\subsection{Semi-Supervised Recognition of Refined Window Structures using Generative Adversarial Networks}
\label{ssec:classification}

Given the window patches and different procedural grammars, the next step is recognizing the specific window type and the corresponding grammar parameters for an image patch.
Unlike the low-level approaches, such as template matching \citep{muller2007imagebased} and line detection \citep{susannebecker2007combined}, we adopt the deep CNNs \citep{he2016deep}.
To facilitate the practical use of our system, we require the number of labeled samples to be two or three magnitudes of order smaller than in previous work \citep{nishida2018procedural}.
In addition, the significantly less manually labeling work enables the joint learning of both types and parameters for the refined window structure.

\subsubsection{Network Architecture}
We build our semi-supervised learning architecture for windows upon DCGAN \citep{radford2016unsupervised} and SGAN \citep{salimans2016improved} for the image generator and semi-supervised learning strategy, respectively.
However, the proposed work also exploits GANs as a global domain adaption tool to transfer the pre-trained weights to fine-grained recognition of window types and parameters, in a separate pre-training stage.
The main difference against DCGAN and SGAN is an additional regression head to estimate the parameters.
For completeness, we also briefly describe the essential aspects of GANs in the following.

\begin{figure}[htbp]
	\centering
	\includegraphics[width=\linewidth]{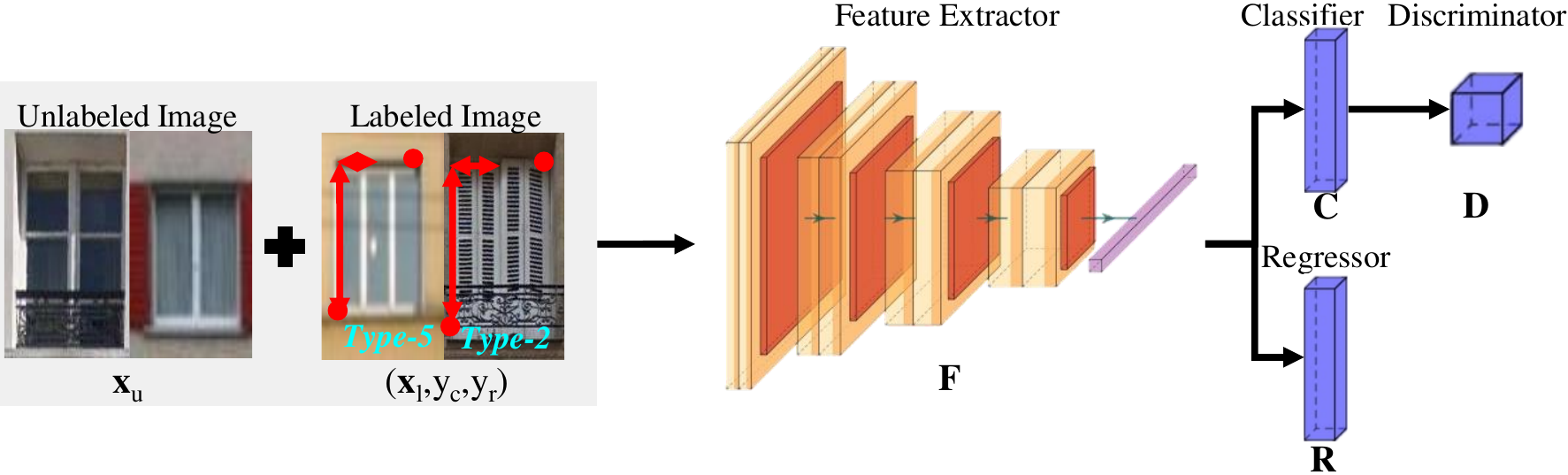}
	\caption{Architecture for semi-supervised fine recognition of window types and parameter estimation of procedural window models.}
	\label{fig:arch}
\end{figure}

A vanilla GAN architecture has two components: the generator $G$ and the discriminator $D$.
The only input is unlabeled data, which generally aims to learn a good generator $G$ for creating realistic data.
The DCGAN \citep{radford2016unsupervised} is a generative model that learns a mapping from a random noise vector $z$ to output an image $\mathbf{x}_g$, $G: z\rightarrow \mathbf{x}_g$.
In contrast, we aim to learn a suitable feature extractor $F$ using a few labeled samples by fine-tuning a pre-trained CNN model.
We intend to solve the domain gap problem with sparse labeled samples and obtain representative features during the fine-tuning: the feature extractor $F$ should be more suitable for downstream tasks, such as classifying the window types and estimating the grammar parameters.

\textbf{Feature Extractor} $F(\mathbf{x})$.
As shown in Figure \ref{fig:arch}, the inputs consist of a small number of labeled window images $\{\mathbf{x}_l, y(y_c,y_r)\}$ and a massive amount of unlabeled window images $\mathbf{x}_u$.
All the images are resized to the same size $\mathbf{x}\in\mathbb{R}^{64\times64}$.
The labels are the window types $y_c$ and the grammar parameters $y_r$ for the window classification and the shape estimation tasks, respectively.
The feature extractor takes an image $\mathbf{x}$ as input and outputs a vector feature $f\in\mathbb{R}^N$ as $f=F(\mathbf{x})$.
The extracted feature $f$ can be used for image classification and parameter regression: we add a linear layer to generate the respective logits for the downstream tasks.
The feature extractor $F$ is agnostic to the adopted CNN architecture, such as ResNet \citep{he2016deep}, VGGNet \citep{simonyan2015very}, EfficientNet \citep{tan2019efficientnet}, \textit{etc}.

\textbf{Classifier} $C(\mathbf{x})$.
When feeding the labeled image $\mathbf{x}_l$ into the pipeline, the linear layer $L_C$ generates a vector of logits $\bm{l}_c=\{l_1,l_2,...,l_K\}\in \mathbb{R}^K$, for $K$ window classes.
Together with the label $y_c$, the logits $\bm{l}_c$ can then be used to train the feature $F$ in a supervised manner.
The probability of the image belonging to the $i$-th window type is computed by applying \textit{softmax}.
\begin{equation}
	p(i)=\frac{\exp(l_i)}{\sum_{k}^{K}\exp(l_k)}
	\label{eq:softmax}
\end{equation}
We term this pipeline as the \textit{classifier} $\bm{l}_c=C(\mathbf{x})\in\mathbb{R}^K$, which intends to infer the window types.

\textbf{Discriminator} $D(\mathbf{x})$.
When feeding the unlabeled $\mathbf{x}_u$ or generated image $\mathbf{x}_g$ into the pipeline, it also first passes through the classifier $C$ and generates $K$-dimensional logits $\bm{l}_c=C(\mathbf{x})$.
Because we only know the real-vs-fake information for $\mathbf{x}_u$ and $\mathbf{x}_g$, the K-D vector $\bm{l}_c$ is not directly usable.
Inspired by \cite{salimans2016improved}, we add a fake class to the original $K$ classes.
Because subtracting any constant from the logit $l_k$ does not change the probability predicted by softmax (Eq. \ref{eq:softmax}), we construct $K+1$ dimensional logits from $\bm{l}_c$ by fixing $l_{K+1}=0$, e.g., $\bm{l}'_c=\{l_1,l_2,...,l_K,0\}$.
In this way, the probability of sample $\mathbf{x}_u$ belongs to the real data (class $i<K+1$) is computed by summing up all the classes using \textit{softmax},
\begin{equation}
	p(i<K+1)=\frac{\sum_{k}^{k=K}\exp(l_k)}{\sum_{k=1}^{K+1}exp(l_k)}=\frac{Z}{Z+1}
	\label{eq:activation}
\end{equation}
where $Z=\sum_{k}^{k=K}\exp(l_k)$.
By the construction above, we obtain an activated scalar value $b$ for binary classification (real-vs-fake), termed as the discriminator $b=D(\mathbf{x})\in\mathbb{R}$ in our architecture.

\textbf{Regressor} $R(\mathbf{x})$.
Another linear layer $L_R$ is attached after the feature extractor $F$, producing a vector of logits $\bm{l}_r\in\mathbb{R}^P$, for $P$ parameters.
The feature $F$ can then be trained jointly with the classifier $C$, using the corresponding parameters $y_r$ in the labeled data.
A set of 6 parameters $\bm{l}_r(\bm{p}_u,\bm{p}_b,\bm{s})$ is considered, including two for the top-left corner $\bm{p}_u$, two for the bottom-right corner $\bm{p}_b$, and two for the cell size $\bm{s}$.
We term the pipeline for parameter estimation as the regressor $\bm{l}_b=R(\mathbf{x})\in\mathbb{R}^P$.
It should be noted that the regressor $R$ follows the global flattened feature from the feature extractor $F$.
Under such architecture, we only have a global sense of the cell size rather than separately estimating the size for each row or column.

\textbf{Generator} $G(z)$.
The same generator as in DCGAN \citep{radford2016unsupervised} is used in our method.
The input is a random vector $\bm{z}\in\mathbb{R}^{100}$.
The first layer projects the random vector to a higher dimension and reshapes the feature to $4\times4\times1024$.
Then, it stacks several deconvolution layers similar to the decoder architecture \citep{springenberg2015striving,long2015fully} and synthesizes a color image $\mathbf{x}_g\in\mathbb{R}^{64\times64\times3}$.
Although the resolution is relatively low compared to the state-of-the-art architecture, 64 pixels in size are good enough for a window patch in practice.
The images $\mathbf{x}_g$ can then be fed into the discriminator pipeline, similar to the unlabeled image $\mathbf{x}_u$.

\subsubsection{Semi-Supervised Adversarial Training}
\label{sssec:semi-training}

\begin{figure}[htb]
	\centering
	\subcaptionbox[a]{Training Discriminator $O_D$}[0.53\linewidth]{\includegraphics[width=\linewidth]{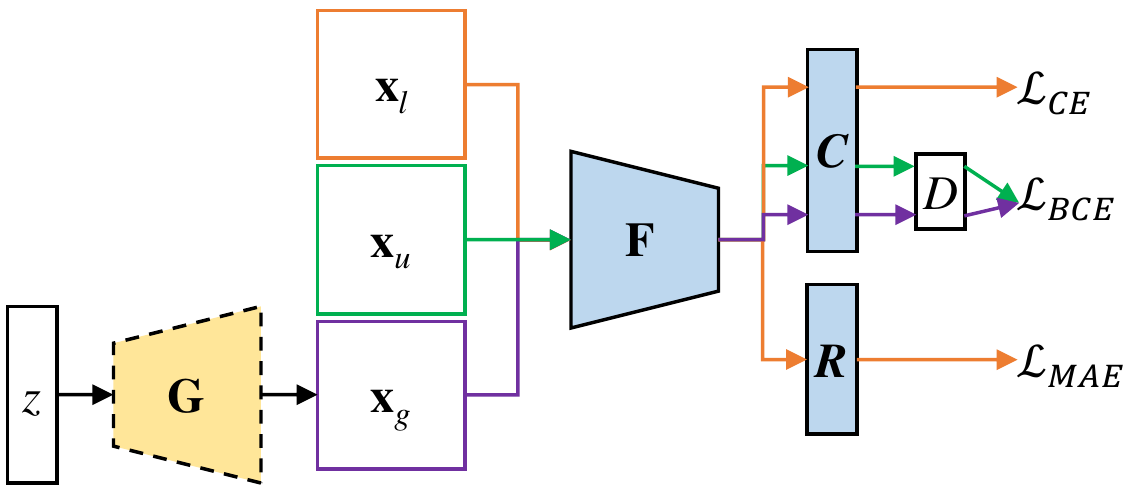}}
	\subcaptionbox[a]{Training Generator $O_G$}[0.45\linewidth]{\includegraphics[width=\linewidth]{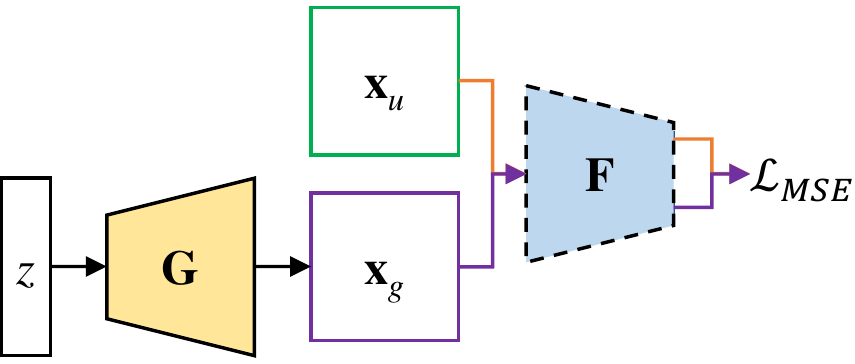}\vspace{1em}}
	\caption{Two optimization stages for the adversarial training. The parts with colored shades denote modules containing learned weights, e.g., generator $G$, feature extractor $F$, linear layers for classifier $C$, and regressor $R$. The dashed boundaries denote fixed modules during training.}
	\label{fig:optimization}
\end{figure}

The learned weights in our architecture consist of a feature extractor $F$, two linear layers ($L_C$ and $L_R$), and a generator $G$ (Figure \ref{fig:optimization}).
The connection between classifier $C$ and discriminator $D$ is just an \textit{ad-hoc} activation function (Eq. \ref{eq:activation}) with no learned weight.
Because the GAN trains the whole network in two separate steps, we set up two optimizers: the training stages for the discriminator $O_D$ and the generator $O_G$.
For $O_D$, weights in $F$, $L_C$, and $L_R$ are back-propagated and adjusted; weights in $G$ are fixed.
The adjusted and fixed weights are the opposite for $O_G$.
The procedural is illustrated in Algorithm \ref{alg:train}.

\begin{algorithm}
	\caption{Semi-Supervised Adversarial Training}\label{alg:train}
	\begin{algorithmic}
		\State Initializing feature extractor $F$ with pre-trained weights.
		\State Initializing generator $G$, classifier $C$ and regressor $R$ randomly.
		\For{each training iteration}
			\State Collecting $m=100$ labeled samples in $(\mathbf{x}_l,y_l)$
			\State Collecting $m$ unlabeled samples in $\mathbf{x}_u$
			\State Generating $m$ random noise $\mathbf{z}$
			\State Adjusting weights $(F,C,R)$ through optimizer $O_D$. \Comment{Equation \ref{eq:lossd}}
			\State Adjusting weights $G$ through optimizer $O_G$. \Comment{Equation \ref{eq:lossg}}
		\EndFor
	\end{algorithmic}
\end{algorithm}

\textbf{Training Discriminator} $O_D$.
For sub-problem $O_D$, the loss function $\mathcal{L}_D$ consists of four parts, 
\begin{equation}
	\mathcal{L}_D=\mathcal{L}_{ce}(C(\mathbf{x}_l),y_c) + \lambda\mathcal{L}_{MAE}(R(\mathbf{x}_l), y_r)+\mathcal{L}_{bce}(D(\mathbf{x}_u), 1)+\mathcal{L}_{bce}(D(G(z)),0),
	\label{eq:lossd}
\end{equation}
where $\mathcal{L}_{ce}$ and $\mathcal{L}_{bce}$ are the standard Cross-Entropy (CE) and Binary Cross-Entropy (BCE) functions, respectively; $\mathcal{L}_{MAE}$ is the mean absolute error function, e.g., the $\mathcal{L}_1$ function.
The window labels supervise the first part $\mathcal{L}_{ce}(C(\mathbf{x}_l),y_c)$.
The window parameters supervise the second part $\mathcal{L}_{MAE}(R(\mathbf{x}_l,y_r)$.
The latter two parts consist of a standard GAN game-value, in which we identify the sampled image without a label $\mathbf{x}_u$ as real (i.e., 1 in the BCE loss) and the generated image $\mathbf{x}_g=G(z)$ as fake (i.e., 0 in the BCE loss).
We add a weight $\lambda$ to balance the loss functions, empirically fixed to $\lambda=2\times10^{-5}$ in our experiments.

\textbf{Training Generator} $O_G$.
In the vanilla GAN, the generator is trained to deceive the discriminator that the generated image is real; accordingly, the loss function is $\mathcal{L}_{bce}(D(G(z)), 1)$.
We take the feature matching loss as described in \cite{salimans2016improved}, in order to better synchronize between training the discriminator and generator, as the following.
\begin{equation}
	\mathcal{L}_G=\mathcal{L}_{MSE}(\mu(F(\mathbf{x}_u)), \mu(F(G(z))) ),
	\label{eq:lossg}
\end{equation}
where $\mathcal{L}_{MSE}$ is the mean squared error loss function (aka. $\mathcal{L}_2$) and $\mu(\cdot)$ is the element-wise average operator in a batch.
Because in each iteration, the training step is fed with different batches of unlabeled images, the feature matching loss can give more diversity, which is essential to prevent G from collapsing to the same content.

\subsubsection{Fine-tuning to Two Separate Stages}
\label{sssec:finetune}

\begin{figure}[htb]
	\centering
	\subcaptionbox[a]{Fine-tuning classifier}[0.48\linewidth]{\includegraphics[width=\linewidth]{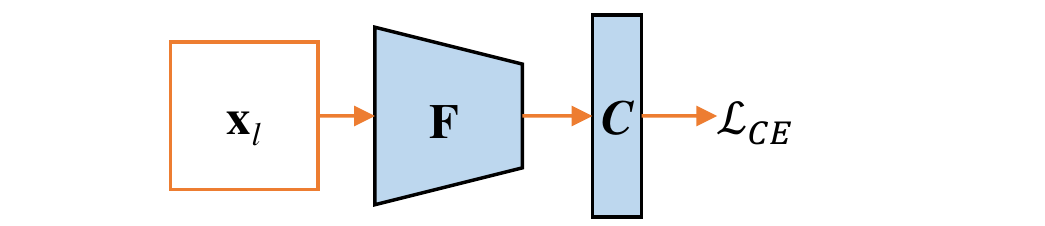}}
	\subcaptionbox[b]{Fine-tuning regressor}[0.48\linewidth]{\includegraphics[width=\linewidth]{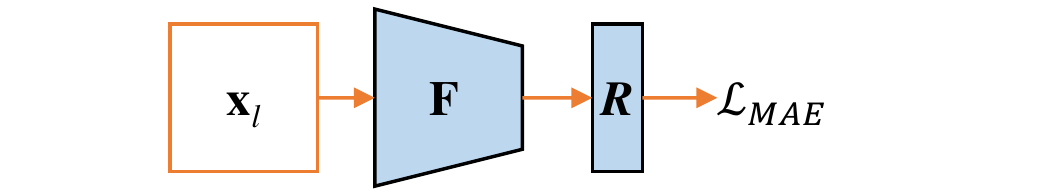}}
	\subcaptionbox[c]{Fine-tuning classifier with adversarial training}[0.48\linewidth]{\includegraphics[width=\linewidth]{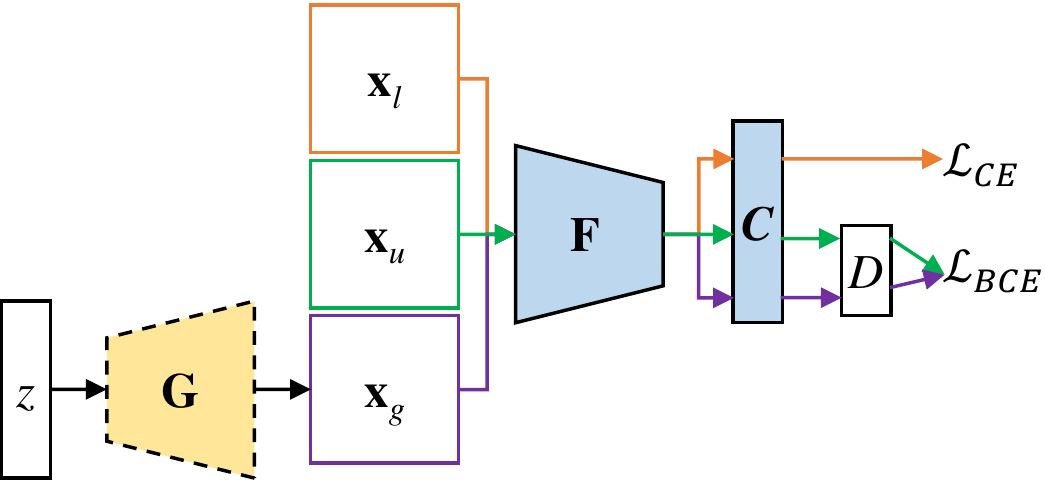}}
	\subcaptionbox[d]{Fine-tuning regressor with adversarial training}[0.48\linewidth]{\includegraphics[width=\linewidth]{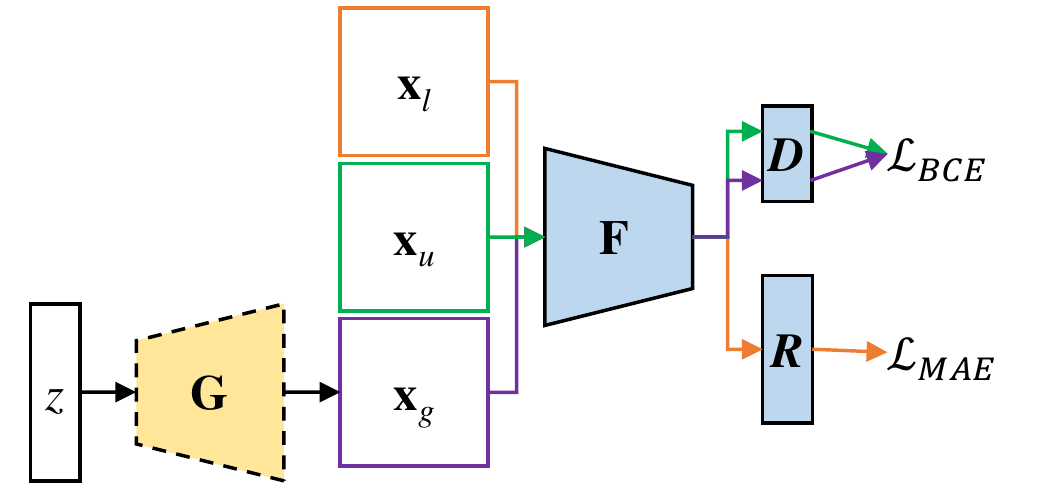}}
	\caption{Fine-tuning of the discriminators for the classification of window types (a,c) and estimation of window parameters (b,d).}
	\label{fig:finetune}
\end{figure}

After pre-training as discussed in Subsection \ref{sssec:semi-training}, the CNN-based feature extractor should be suitable for tasks of the window domain.
Thanks to the GAN-based semi-supervised learning, the pre-training is already fed with abundant unlabeled window patches, improving the generalization capability for different window styles or image sources.
The learned feature extractor $F$ above can simultaneously classify window types and estimate the procedural parameters.
However, slightly modifying the architecture and fine-tuning the weights to fit only a single purpose will naturally improve performance. 
The regressor $R$ especially needs further fine-tuning because the discriminator $D$ is actually biased to  the classifier, because the discriminator is essentially a weak classifier between real and fake windows.
The only drawback is that, at testing time, two forward passes are required.
With modern computing units, the overhead caused by this issue is almost negligible in our whole pipeline.
Therefore, we transfer the weights into two separate models for the classification and regression problems.

We coin two fine-tuning strategies, as shown in Figure \ref{fig:finetune}.
The first is the classical supervised training strategy by adjusting the feature extractor and the specific classification or regression heads (Figure \ref{fig:finetune} top row).
The second fine-tuning strategy is similar to Section \ref{sssec:semi-training} (Figure \ref{fig:finetune} bottom row).
The differences only lie in the discriminator $O_D$.
When fine-tuning the classifier $C$, we omit the regressor branch $R$; the cross-entropy loss $\mathcal{L}_{CE}$ for the window types and two binary cross-entropy losses $\mathcal{L}_{BCE}$ for the real-vs-fake GAN game are used.
On the contrary, when fine-tuning the regressor $R$, we omit the classifier branch $C$ and add the classical discriminator layer $D$; the two binary cross-entropy losses $\mathcal{L}_{BCE}$ and the mean absolute error $\mathcal{L}_{MAE}$ for regression are used.
After fine-tuning, the classification (Figure \ref{fig:finetune}a) and regression heads (\ref{fig:finetune}b) are exploited in the inference.

\subsection{Interactive Modeling System with Window Instances}
\label{ssec:instancing}

\begin{figure}[htb]
	\centering
	\includegraphics[width=\linewidth]{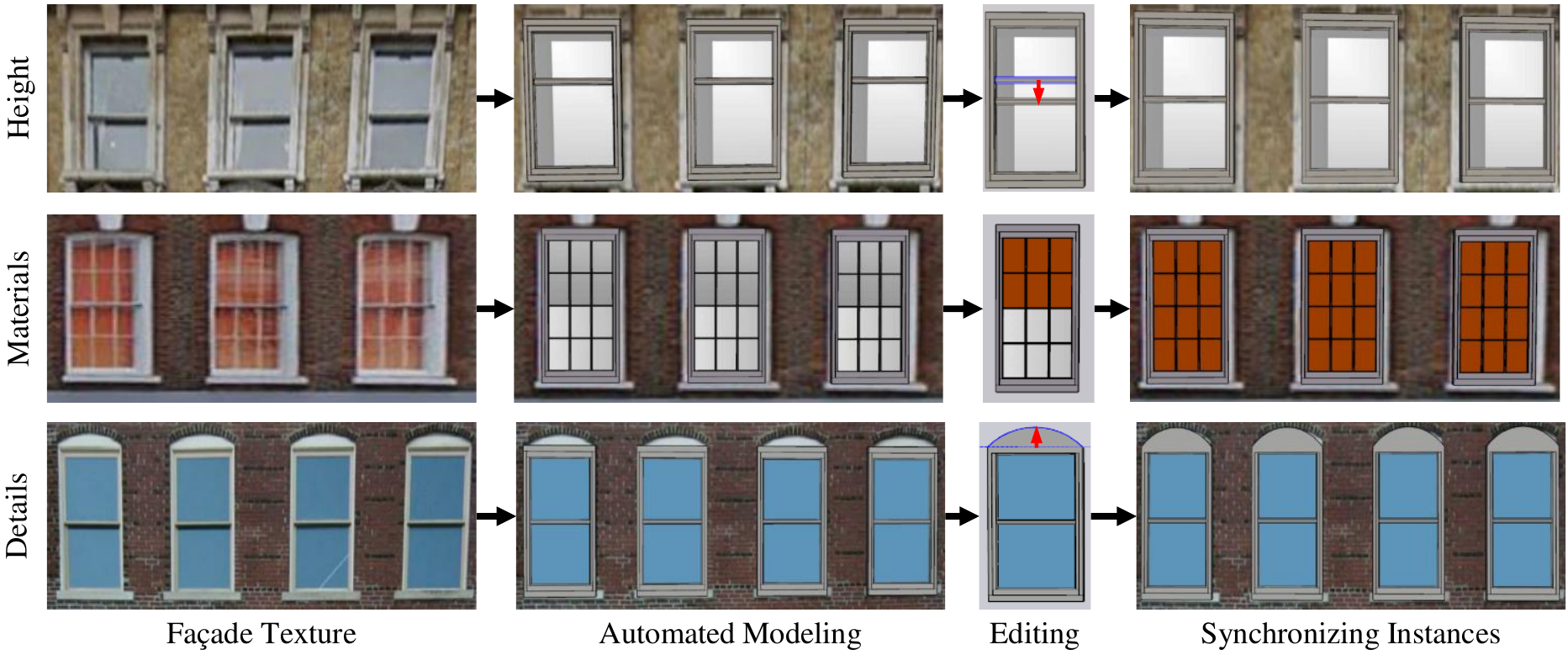}
	\caption{Typical interactive editing operations and batch synchronization of the same component instance.}
	\label{fig:edit}
\end{figure}

As shown in Figure \ref{fig:edit}, we may encounter numerous undesired situations in real-world applications, impeding a fully automatic pipeline.
The distortions on window textures will cause incorrect parameters (top row).
Window materials that are different from the default will influence visual reality (middle row).
Various geometry details not considered in the procedure grammar will cause misalignment between window models and textures (bottom row).
The proposed prototype incorporates several interactive editing steps into the following three stages.

\paragraph{1) Marking of windows}
As shown in Figure \ref{fig:overview}, after detecting the window bounding boxes, we allow interactive addition and cleaning of the windows to improve the detection quality, as a preprocessing step.
In addition, a set of suitable window bounding boxes $\{\mathcal{B}_i\}$ is marked as the same cluster, and a single geometry instance $\mathcal{C}_k$ is created for a window cluster.
Currently, we only consider interactive clustering of windows using the selection tool in the modeling software.

\paragraph{2) Averaging inference results}
At testing time, we infer the window type through the classifier and parameters through the regressor for each image patch.
A procedure grammar $\mathcal{G}_i$ is assembled accordingly.
Results for windows belonging to the same cluster are averaged to improve accuracy and regularity.
The mode of the window types is taken for the grammar of the instance cluster $\mathcal{C}_k$. 
The mesh models of the geometric instances are then created using the averaged grammar.

\paragraph{3) Instance editing and batch synchronization}
Multiple meshes sharing the same instanced cluster $\mathcal{C}_k$ can be efficiently generated by applying different transformation matrices using the modeling API \citep{trimble2021sketchup}.
Because the underlying geometries are shared through different windows, the modifications to the geometric instances will simultaneously synchronize to all the models in the same cluster, as shown in Figure \ref{fig:edit}.

\section{Experimental evaluations and analyses}
\label{sec:experiments}
\subsection{Dataset and implementation details}

\subsubsection{Dataset description}

In our experiments, we adopted the publicly available fa\c{c}ade dataset for evaluation.
Specifically, the CMP fa\c{c}ade \citep{tylevcek2013spatial} and LSAA (Large Scale Architectural Asset) datasets \citep{zhu2020large} were used.
The CMP fa\c{c}ade dataset includes 378 labeled images with 12 semantic classes.
We only kept the parts of windows through interactively cropping in LabelImg \citep{lin2015labelimg}.
In this way, about $10^4$ window patches were obtained.
The LSAA dataset provides more than $10^5$ fa\c{c}ade images containing semantics of windows, doors, and balconies.
We cropped the windows with the provided bounding boxes.
In summary, we obtained about $1.6\times10^5$ window patches from the above two datasets.

It should be noted that the above window patches were not associated with fine-grained types and parameters.
In addition, we intentionally conducted a small amount of dilation to the bounding boxes of windows during the cropping.
For each type in Figure \ref{fig:types}, we made 1000 labeled samples for both training and testing, following \cite{deng2009imagenet}.
The parameters in Figure \ref{fig:types} of the windows were also obtained in LabelImg as two bounding boxes, e.g., one for the two corners and one for the cell size.
We only chosen 100 images per category for training to simulate the real-world applications, and the rests were used for testing.
The same training dataset was used for both the classification and regression problems; thus, 900 labeled and $1.6\times10^5$ unlabeled image patches were used in the following experiments.
In the pre-training stage, we trained a single multi-task network for the classifier $C$ and the regressor $R$.
In the following fine-tuning stage, the multi-task network was separated into two networks, one for the classifier $C$ and one for the regressor $R$, as shown in Figure \ref{fig:finetune}.

For the 3D modeling, the inputs to our system are textured fac\c{a}des and pre-processed window semantics.
In this regard, publicly available crowd-sourcing datasets \citep{sketchup20213d} and free samples from the 3D model vendors \citep{accucities20213d} are used.
In addition, datasets collected by the authors are also adopted to increase the variety of the models.
In preprocessing, the bounding boxes of the windows were detected using a pre-trained faster-RCNN as in our previous work \citep{hu2020fast} and exported as the LabelImg format.
If necessary, we double-checked the detection results and interactively polished the results by adding occluded windows and removing false detection.
Detailed information of the models is included in Table \ref{tab:dataset}, in which we list the sources of textures, approximated resolutions, locations, and styles of the models.
Because the training samples mainly cover building styles in the Europe, datasets covering London were used for detailed evaluations; nevertheless, models in other cities were also considered to evaluate the generalization ability of the proposed method.
The preprocessed datasets will be made available on the authors' project page\footnote{\url{https://vrlab.org.cn/~hanhu/projects/windows}}.

\begin{table}[htb]
	\caption{Textured models used for fa\c{c}ade modeling.}
	\label{tab:dataset}
	\resizebox{\textwidth}{!}{%
		\begin{tabular}{@{}llllll@{}}
			\toprule
			Name                            & Model Source       & Texture Source & GSD ($cm$) & Coverage   & City       \\ \midrule
			Bank                            & UAV Photogrammetry & UAV            & 2          & Individual & Huzhou, CN \\
			Central Business District (CBD) & 3D Warehouse       & Aerial         & 10         & Individual & London, UK \\
			State Capitol                   & 3D Warehouse       & Ground         & 2          & Individual & Denver, US \\
			Borough High Street             & AccuCities         & Ground         & 2          & Block      & London, UK \\
			Broad Street                    & 3D Warehouse       & UAV            & 5          & Block      & Rome, US  \\ \bottomrule
		\end{tabular}%
	}
\end{table}

\subsubsection{Implementation details}
We implemented the recognition parts in PyTorch.
The Adam optimizer was used in all the experiments, with default hyperparameters.
We trained all the models for relatively large epochs (300 in this paper) because of the small number of labeled samples.
In fact, the unlabeled data will be merely trained for five epochs considering the unbalanced numbers between labeled and unlabeled images.
As noted by \cite{salimans2016improved}, batch training is essential for GAN to converge.
The batch size was fixed to $m=100$ (as an exact division of the labeled samples) for all the labeled, unlabeled, and generated samples.
The whole training procedure took about two hours on a single NVIDIA RTX 3080 graphics card.

\subsection{Experimental results for window recognition}
\subsubsection{Comparisons on classification accuracy and parameter regression}
This section evaluates three different models, including two classic networks, i.e., VGG \citep{simonyan2015very} and ResNet \citep{he2016deep}, and one recent state-of-the-art, i.e., EfficientNet \citep{tan2019efficientnet}.
Because the architecture is essentially the same, we denote the proposed approach with a GAN suffix to each network.
It should be noted that we use a variant of VGG model with batch normalization; otherwise, the semi-supervised training with GAN will not converge.

Three top-accuracies and the average regression loss value ($\mathscr{L}_{MAE}$ in Equation \ref{eq:lossd}) are reported in Table \ref{tab:recognition}.
For the classification of window types, the proposed adversarial training strategy outperforms the supervised counterpart by a clear margin (3\% to 7\%) in top-1 accuracy, which reveals the ability for automatic recognition.
This trend is consistent with top-2 and top-3 accuracies, corresponding to the performances with interactively user selection.
Turning to the regression accuracy (last column in Table \ref{tab:recognition}), the improvements in the localization and the size estimation are also quite clear, about 30\% to 70\% relative increases.
Another finding is that the performances in both the classification and the regression across all the networks are more stable with the proposed strategy; this is also quite important in real-world applications.

\begin{table}[htb]
	\centering
	\caption{Comparison between semi-supervised training (denoted as GAN) and vanilla supervised training. The symbol $\uparrow$ indicates higher is better and $\downarrow$ for lower. The \textit{Rel.} column denotes the improvements between GAN and vanilla model.}
	\label{tab:recognition}
	\resizebox{\textwidth}{!}{%
		\begin{tabular}{@{}cccccccccc@{}}
			\toprule
			\multirow{2}{*}{Network} &
			\multirow{2}{*}{Variant} &
			\multicolumn{2}{c}{Top-1 (\%) $\uparrow$} &
			\multicolumn{2}{c}{Top-2 (\%) $\uparrow$} &
			\multicolumn{2}{c}{Top-3 (\%) $\uparrow$}&
			\multicolumn{2}{c}{$\mathscr{L}_{MAE}$ (pixel) $\downarrow$} \\
			&         & Acc.  & Rel. & Acc.  & Rel. & Acc.  & Rel. & Val.  & Rel.   \\ \midrule
			\multirow{2}{*}{VGG-19}          & GAN     & 79.25 & 7.02 & 90.11 & 3.31 & 93.91 & 1.82 & 21.04 & -16.22 \\
			& Vanilla & 72.23 & /    & 86.8  & /    & 92.09 & /    & 37.26 & /      \\ \midrule
			\multirow{2}{*}{ResNet-34}       & GAN     & 75.54 & 7.8  & 88.72 & 6.09 & 93.31 & 3.83 & 24.90 & -11.72 \\
			& Vanilla & 67.74 & /    & 82.63 & /    & 89.48 & /    & 36.62 & /      \\ \midrule
			\multirow{2}{*}{EfficientNet-B0} & GAN     & 74.57 & 3.22 & 87.59 & 0.7  & 93.04 & 0.46 & 22.66 & -6.45  \\
			& Vanilla & 71.35 & /    & 86.89 & /    & 92.58 & /    & 29.11 & /      \\ \bottomrule
		\end{tabular}%
	}
\end{table}

\begin{figure}[H]
	\centering
	\includegraphics[width=\linewidth]{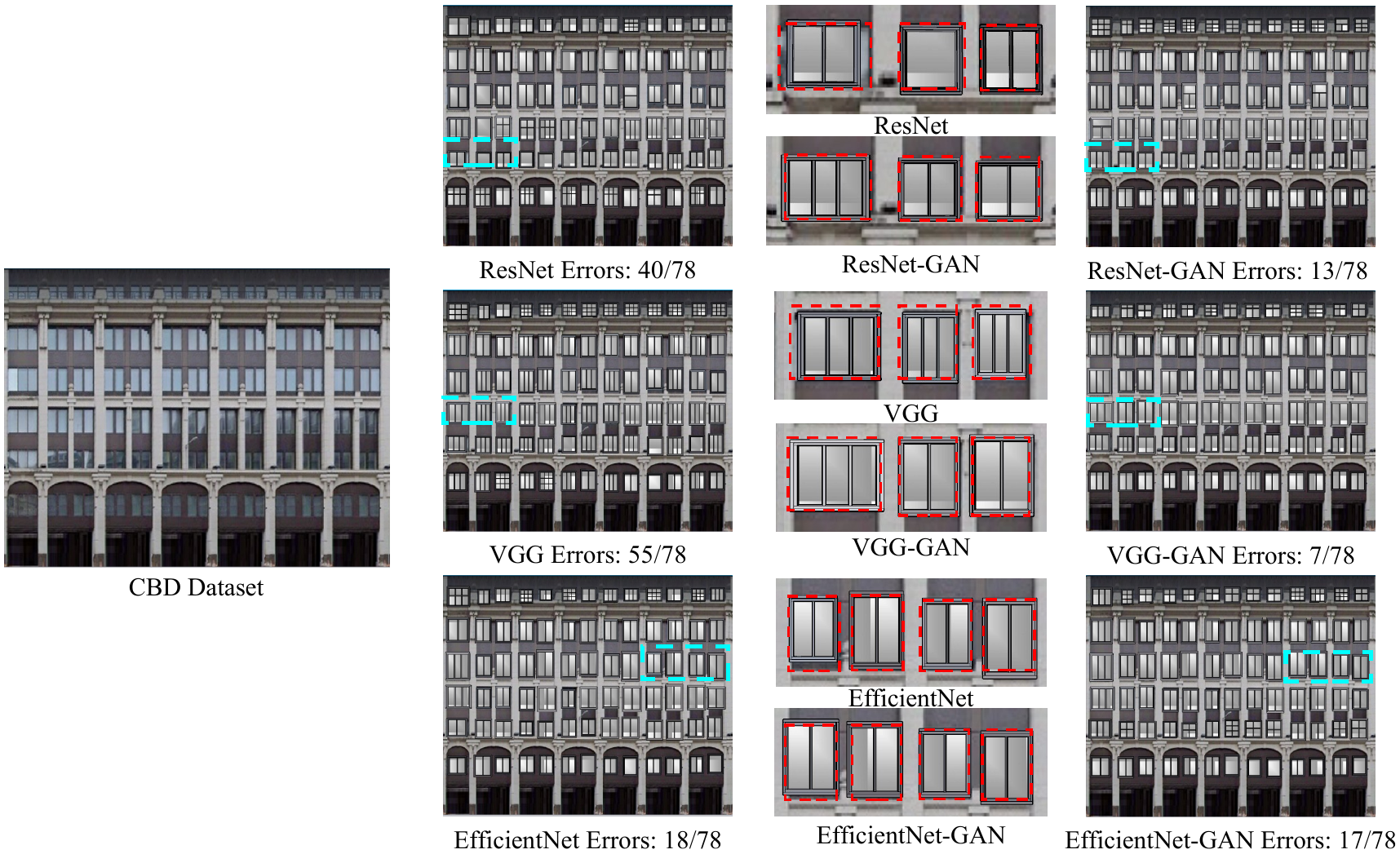}
	\caption{Qualitative comparison of different networks using texture images in the CBD dataset. The second and fourth columns show automatic recognition results overlaid on the image. The third column demonstrates an enlarged area highlighted by the cyan boxes. The red boxes indicate the ground truth window regions.}
	\label{fig:qualitative_compare_cbd}
\end{figure}

\begin{figure}[H]
	\centering
	\includegraphics[width=\linewidth]{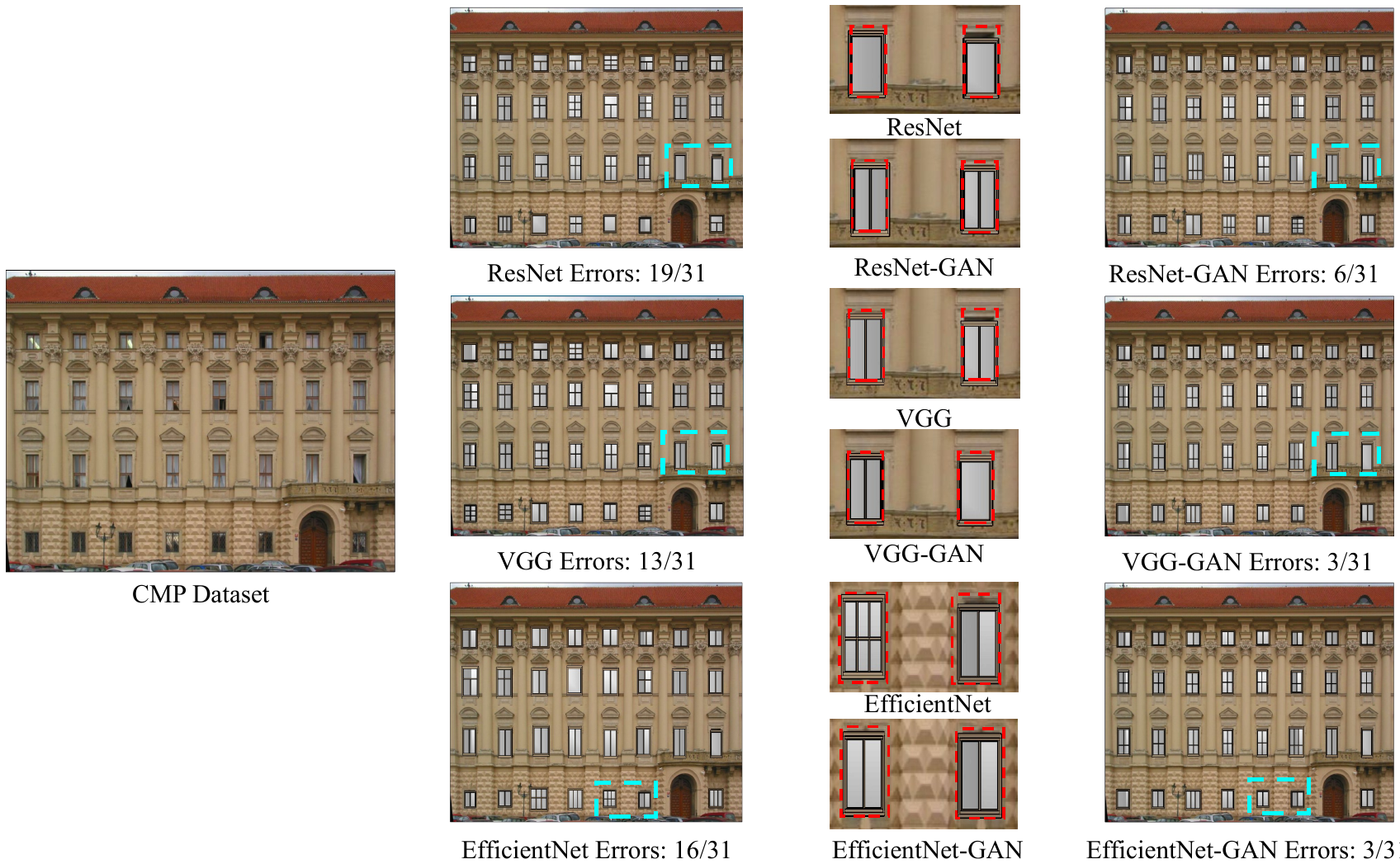}
	\caption{Qualitative comparison of different networks using a typical image in CMP fa\c{c}ade dataset. The second and fourth columns show automatic recognition results overlaid on the image. The third column demonstrates an enlarged area highlighted by the cyan boxes. The red boxes indicate the ground truth window regions.}
	\label{fig:qualitative_compare_cmp}
\end{figure}

Two fa\c{c}ade images were chosen to evaluate the proposed method qualitatively, as shown in Figures \ref{fig:qualitative_compare_cbd} and \ref{fig:qualitative_compare_cmp}.
The first (CBD dataset) features reflective glassy windows, and the second in front of the curtains.
The texture images are shown in the left column in each figure.
The second and fourth columns demonstrate results obtained without and with the adversarial training strategy, respectively.
The third column presents an enlarged subset indicated by the cyan rectangles.
For VGG and ResNet, almost half the windows are classified incorrectly.
And the parameters are also less accurate without the proposed method, as indicated by the red boxes in the enlarged views.

\subsubsection{Ablation studies}
\label{sssec:ablation}
The proposed method aims to alleviate the domain gap between ImageNet pre-training results and the fine-grained recognition for the windows dataset under only sparse labeled samples.
We conducted two ablation studies to validate the proposed method for classification and regression, as shown in Tables \ref{tab:ablation-class} and \ref{tab:ablation-reg}, respectively.
The classical fine-tuning strategy using pre-trained weights from ImageNet is considered as baseline, e.g., the fourth row in each table.
The semi-supervised pre-training as described in Subsection \ref{sssec:semi-training} tries to transfer the weights from the domain of ImageNet to windows, as the first row in each table (ResNetCR+GAN).
The resulted weights of ResNetCR+GAN are used as pre-trained weights for further fine-tuning (the \textit{Window} column in Tables \ref{tab:ablation-class} and \ref{tab:ablation-reg}).
Then two different strategies fine-tune the feature extractor of ResNetCR+GAN for classification and regression problem separately, as shown in the second and third rows of Tables \ref{tab:ablation-class} and \ref{tab:ablation-reg}.

\begin{table}[htb]
	\centering
	\caption{Evaluation for the classification accuracy with different training strategies and ResNet as backbones. For the strategy names, suffix \textit{CR} indicates training for both classifiers and regressors, \textit{GAN} for semi-supervised training strategy using unlabeled data, \textit{WP} for using semi-supervised windows dataset as pre-training weights, e.g., results of ResNetCR+GAN, and \textit{NP} for using no pre-trained weights.}
	\label{tab:ablation-class}
	\resizebox{\textwidth}{!}{%
		\begin{tabular}{lcccccc}
			\hline
			\multirow{2}{*}{Name} & \multicolumn{2}{c}{Labeled Data} & Unlabeled  & \multicolumn{2}{c}{Pre-trained Weights} & \multirow{2}{*}{Top-1 Accurary} \\
			& Classes         & Parameters     & Data       & ImageNet         & Window      &                                 \\ \hline
			ResNetCR+GAN          & \checkmark      & \checkmark     & \checkmark & \checkmark       &                   & 75.54                           \\
			ResNet+GAN+WP         & \checkmark      &                & \checkmark &                  & \checkmark        & 76.14                           \\
			ResNet+WP             & \checkmark      &                &            &                  & \checkmark        & 78.91                           \\
			ResNet                & \checkmark      &                &            & \checkmark       &                   & 67.74                           \\ \hline
			ResNet+NP             & \checkmark      &                &            &                  &                   & 38.75                           \\
			ResNet+GAN+NP         & \checkmark      &                & \checkmark &                  &                   & 61.25                           \\ \hline
		\end{tabular}%
	}
\end{table}

\begin{table}[htb]
	\centering
	\caption{Evaluation for the regression accuracy with different training strategies and ResNet as backbones. The symbols are the same with Table \ref{tab:ablation-class}.}
	\label{tab:ablation-reg}
	\resizebox{\textwidth}{!}{%
		\begin{tabular}{lcccccc}
			\hline
			\multirow{2}{*}{Name} & \multicolumn{2}{c}{Labeled Data} & Unlabeled  & \multicolumn{2}{c}{Pre-trained Weights} & \multirow{2}{*}{$\mathscr{L}_{MAE}$} \\
			& Classes         & Parameters     & Data       & ImageNet         & Window      &                                      \\ \hline
			ResNetCR+GAN          & \checkmark      & \checkmark     & \checkmark & \checkmark       &                   & 24.90                                \\
			ResNet+GAN+WP         & \checkmark      &                & \checkmark &                  & \checkmark        & 19.91                                \\
			ResNet+WP             & \checkmark      &                &            &                  & \checkmark        & 17.88                                \\
			ResNet                & \checkmark      &                &            & \checkmark       &                   & 36.62                                \\ \hline
			ResNet+NP             & \checkmark      &                &            &                  &                   & 29.57                                \\
			ResNet+GAN+NP         & \checkmark      &                & \checkmark &                  &                   & 20.09                                \\ \hline
		\end{tabular}%
	}
\end{table}

It could be noted that even trained for two purposes, ResNetCR+GAN still outperforms the baseline by a large margin.
In addition, after fine-tuning the pre-trained weights above, the performances also increase for both tasks.
The training settings between ResNet+WP and baseline, i.e., the third and fourth rows, are almost the same: they use the same architecture and data.
The only difference is the pre-trained weight, ResNet from the ImageNet domain and ResNet+WP in the window domain.
Just using different pre-trained weights leads to about an 11\% leap in the classification of window types and about 1-fold increases for the parameter estimation of the window grammars.
This is because the pre-training stage (ResNetCR+GAN) already obtains a model suitable for the window domain, i.e., transferring the domain from ImageNet to the windows.
Further fine-tuning using a small number of labeled samples in the same domain can also slightly improve the performances.

It is also interesting to compare the performances of fine-tuning a window pre-trained weights with and without unlabeled data (ResBet+GAN+WP and ResNet+WP).
They both target a single task and demonstrate improvements to the multi-task ResNetCR+GAN.
ResNet+WP demonstrates slightly better performances in both Tables \ref{tab:ablation-class} and \ref{tab:ablation-reg}.
This is probably because that the initial weights are already in the window domain.
In addition, the adversarial strategy can be considered as a regularization term pulling the model away from the perfect fitting of the data.

We also compare results with no pre-training weight, i.e., the last two rows in Tables \ref{tab:ablation-class} and \ref{tab:ablation-reg}.
For classification, the performances with no pre-trained weight are significantly inferior to the variants with the pre-trained weights.
For regression, we can still obtain satisfactory results using the semi-supervised training strategy even with no pre-trained weight; in addition, training from scratch even outperforms fine-tuning from ImageNet weights, i.e., ResNet vs. ResNet+NP in Table \ref{tab:ablation-reg}.
This is probably because weights from ImageNet were initially used for classification purposes, making the fine-tuning for regression difficult.
Nonetheless, the training with additional unlabeled data is always more stable than the counter-parts with only labeled images.

\subsection{Experimental results for 3D fa\c{c}ade modeling}
\subsubsection{Effects of grouped inference}
\label{sssec:grouped}

For a practical solution in 3D modeling of the fa\c{c}ade, interactions are somehow inevitable, considering the limitation in accuracies, various fa\c{c}ade styles, and regularity of the buildings.
As described in Section \ref{ssec:instancing}, a simple yet effective approach to improve recognition accuracy and modeling regularity is the interactively grouping of windows and batch voting of the results.

To validate the efficacy of the interactive grouping, we chose two blocks of buildings, e.g., Borough High Street and Broad Street.
As comparison, we chose automatic results (e.g. Top-1) inferred from both ResNet and ResNet+GAN as the baselines.
Results are illustrated in Table \ref{tab:group_inference}.

\begin{table}[htb]
	\centering
	\caption{Quantitative evaluations of interactively grouped inference on the two street blocks. GAN denotes adversarial training and GI for grouped inference. The relative improvement (Rel.) is tested against the variant without GI.}
	\label{tab:group_inference}
	\resizebox{\textwidth}{!}{%
		\begin{tabular}{@{}llllllllll@{}}
			\toprule
			\multirow{2}{*}{Dataset} &
			\multirow{2}{*}{Variant Name} &
			\multicolumn{2}{l}{Top-1 (\%) $\uparrow$} &
			\multicolumn{2}{l}{Top-2 (\%) $\uparrow$} &
			\multicolumn{2}{l}{Top-3 (\%) $\uparrow$} &
			\multicolumn{2}{l}{$\mathscr{L}_{MAE}$ (pixels) $\downarrow$} \\
			&               & Acc.  & Rel.  & Acc.  & Rel. & Acc.  & Rel. & Val.  & Rel.  \\ \midrule
			\multirow{4}{*}{Borough High Street} & ResNet+GAN+GI & 82.89 & 12.17 & 89.35 & 9.88 & 93.92 & 8.75 & 24.53 & -0.64 \\
			& ResNet+GAN    & 70.72 & /     & 79.47 & /    & 85.17 & /    & 25.17 & /     \\
			& ResNet+GI     & 53.99 & 8.74  & 63.88 & 1.90 & 74.14 & 3.80 & 32.72 & -1.44 \\
			& Resnet        & 45.25 & /     & 61.98 & /    & 70.34 & /    & 34.16 & /     \\
			\multirow{4}{*}{Broad Street}        & ResNet+GAN+GI & 81.94 & 9.72  & 91.67 & 5.56 & 95.83 & 5.55 & 24.70 & -0.34 \\
			& ResNet+GAN    & 72.22 & /     & 86.11 & /    & 90.28 & /    & 25.04 & /     \\
			& ResNet+GI     & 50.00 & 8.33  & 63.80 & 2.69 & 73.61 & 0.00 & 31.87 & -1.36 \\
			& Resnet        & 41.67 & /     & 61.11 & /    & 73.61 & /    & 33.23 & /     \\ \bottomrule
		\end{tabular}%
	}
\end{table}

It can be noticed that the voting strategy, although almost as simple as plain averaging, can significantly increase the performances to about 10\% in Top-1 accuracy.
In addition, the performance boost in the classification of window types for a better baseline (ResNet+GAN) is even more significant than the inferior one (ResNet).
This seems counter-intuitive since improving an already good result is much more challenging.
However, good results will get even better in voting, as long as the performance for a single vote is beyond random guess.
Although the enhancements for parameter regression are less impressive, we will find later in the qualitative results that the regularities are better.

\begin{figure}[H]
	\centering
	\includegraphics[width=\linewidth]{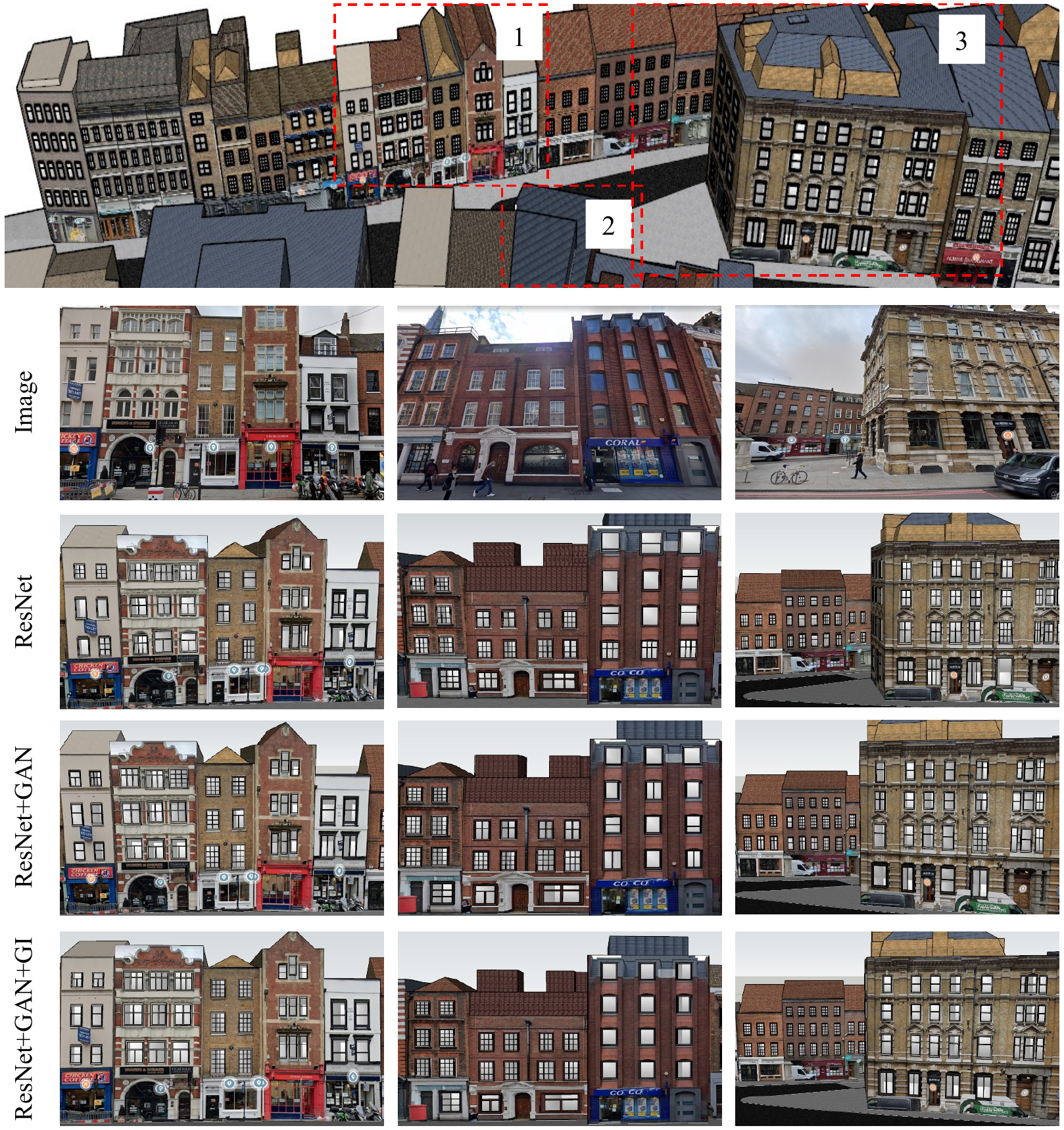}
	\caption{Comparison of different inference strategies in the Borough High Street scene. The red boxes and numbers indicate the enlarged areas below and the columns, respectively.}
	\label{fig:model_borough}
\end{figure}

\begin{figure}[H]
	\centering
	\includegraphics[width=\linewidth]{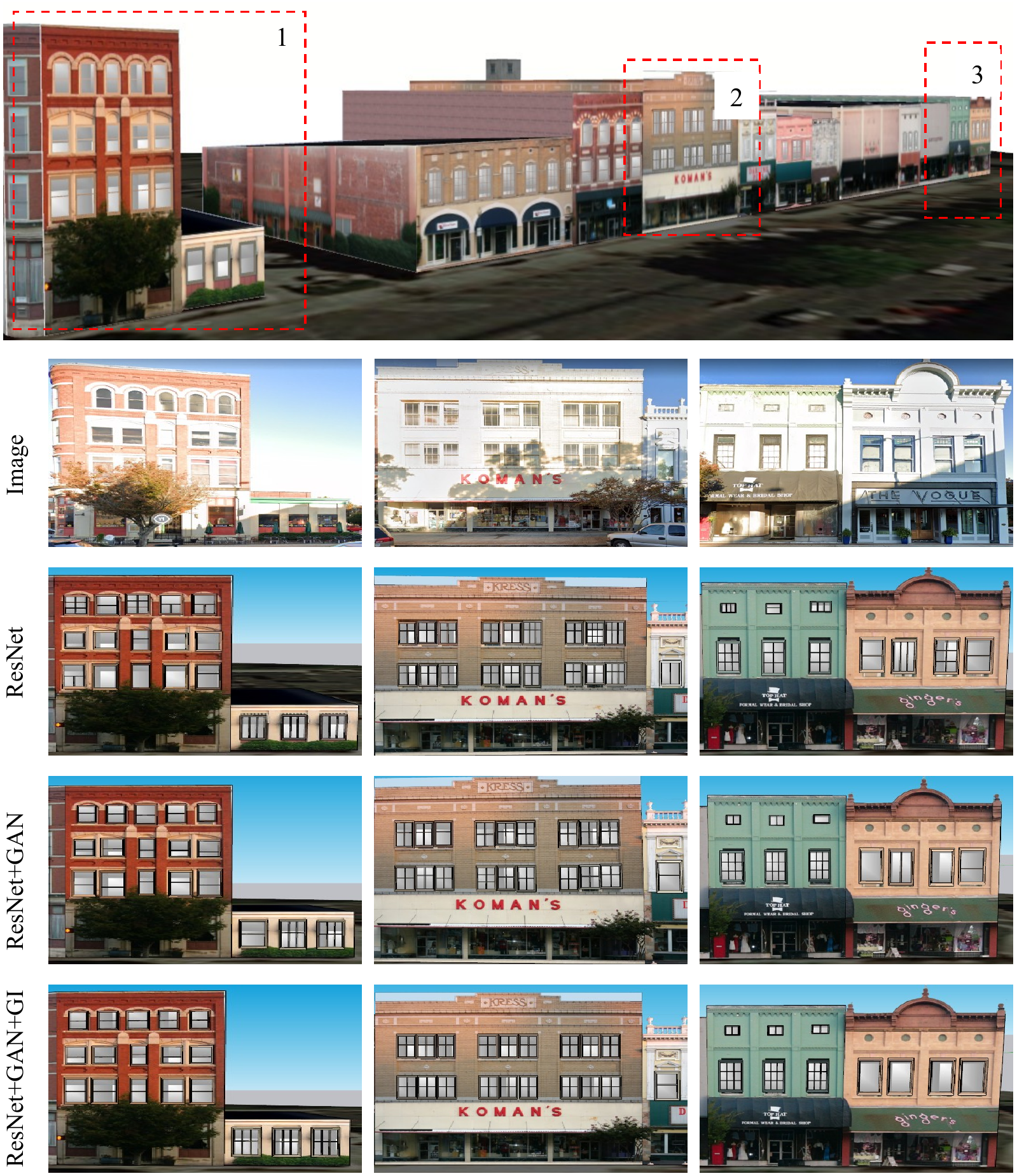}
	\caption{Comparison of different inference strategies in the Broad Street scene. The red boxes and numbers indicate the enlarged areas below and the columns, respectively.}
	\label{fig:model_broad}
\end{figure}

Figures \ref{fig:model_borough} and \ref{fig:model_broad} show the 3D models with enlarged views in the Borough High Street and Broad Street scenes.
It can be noted that with just simple interactions of selecting groups, the regularities of the 3D models have been significantly improved.
The improved regularities are revealed in two aspects, i.e., the types and alignments.
In most cases, enforcing regularity in window types will also increase the accuracy of classification, which is consistent with Table \ref{tab:group_inference}.
Although the quantitative improvement in parameter estimation is less significant, the visual quality is much more appealing, such as the first column of the Borough High Street scene.

\begin{figure}[htp]
	\centering
	\includegraphics[width=\linewidth]{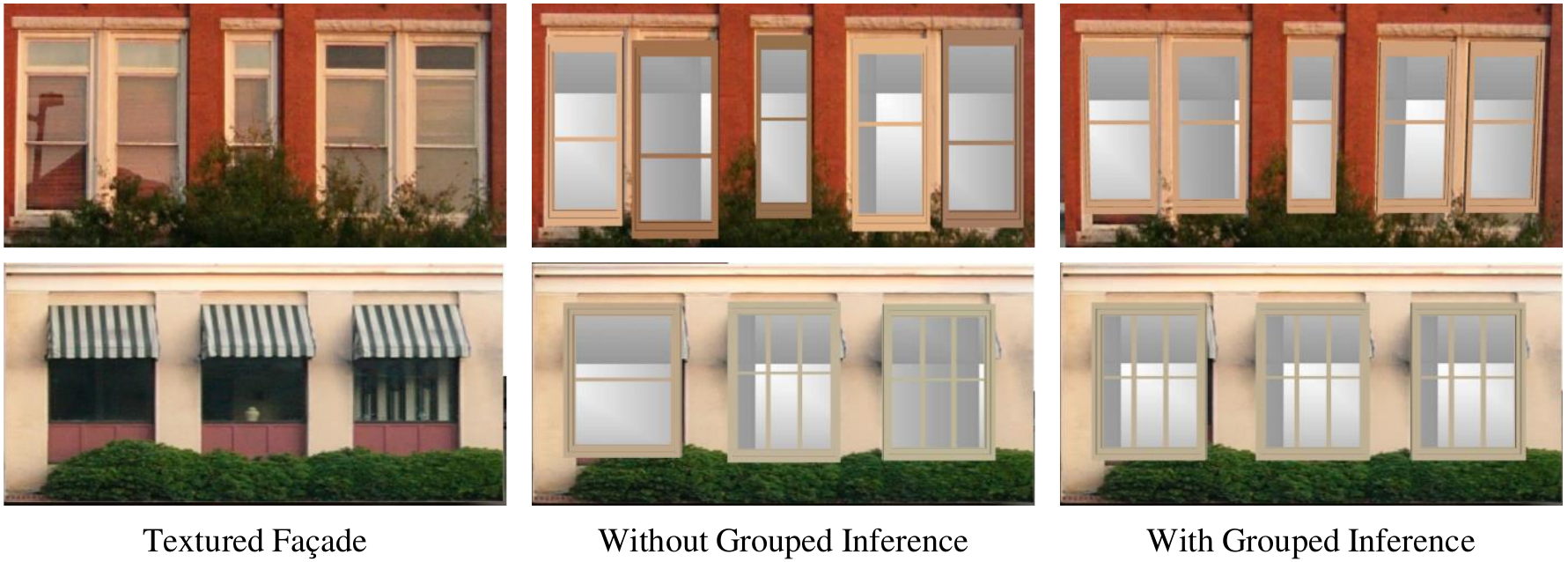}
	\caption{Comparisons with and without grouped inferencing for partially occluded fa\c{c}ades.}
	\label{fig:occluded}
\end{figure}

Another interesting effect of the grouped inferencing is to improve the performances for partially occluded windows, commonly seen in street-view and aerial oblique images, as shown in Figure \ref{fig:occluded}.
Without grouped inference (middle column), the classification and regression accuracies are not satisfactory.
But after clustering suitable windows into the corresponding groups interactively, the results are better aligned with human expectations.

\subsubsection{Results of 3D models}
\begin{figure}[htb]
	\centering
	\includegraphics[width=\linewidth]{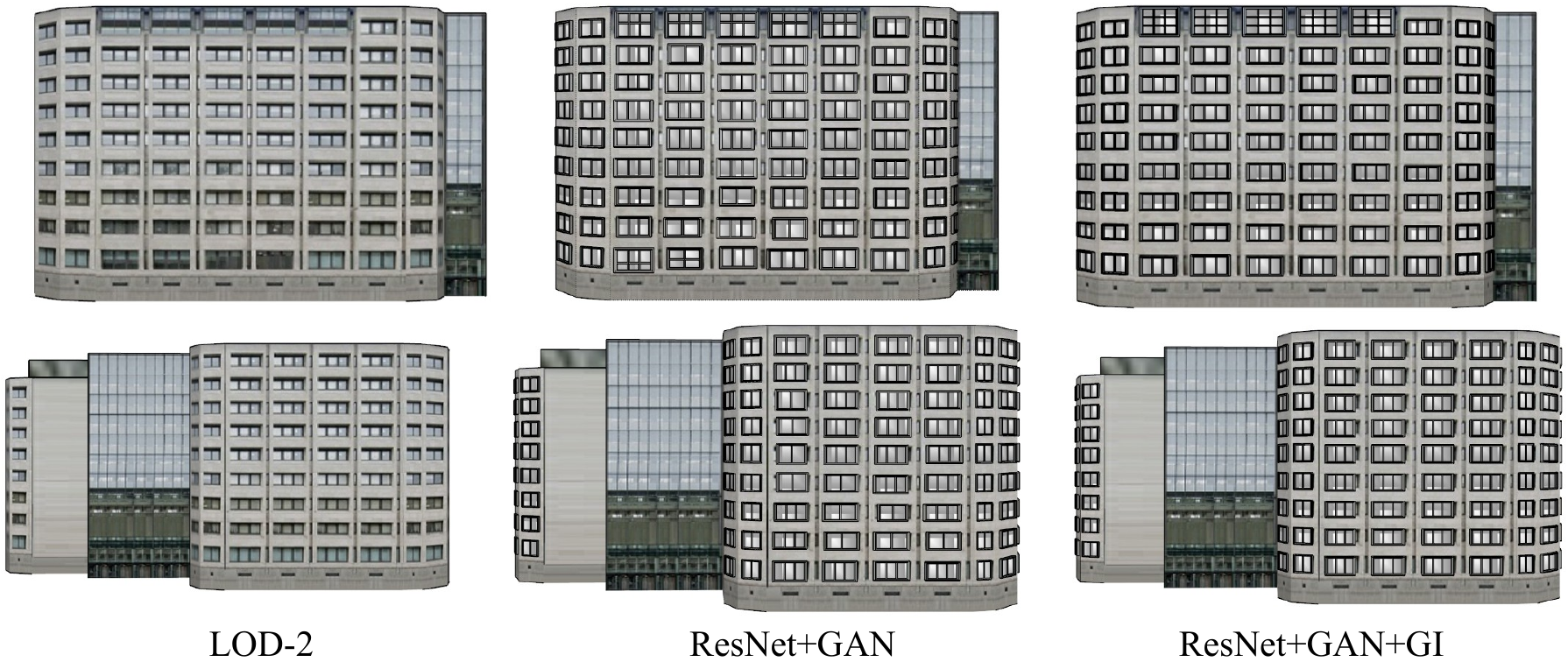}
	\caption{Reconstruction results for the BT Center in the CBD dataset.}
	\label{fig:models_btcenter}
\end{figure}

\begin{figure}[htb]
	\centering
	\includegraphics[width=\linewidth]{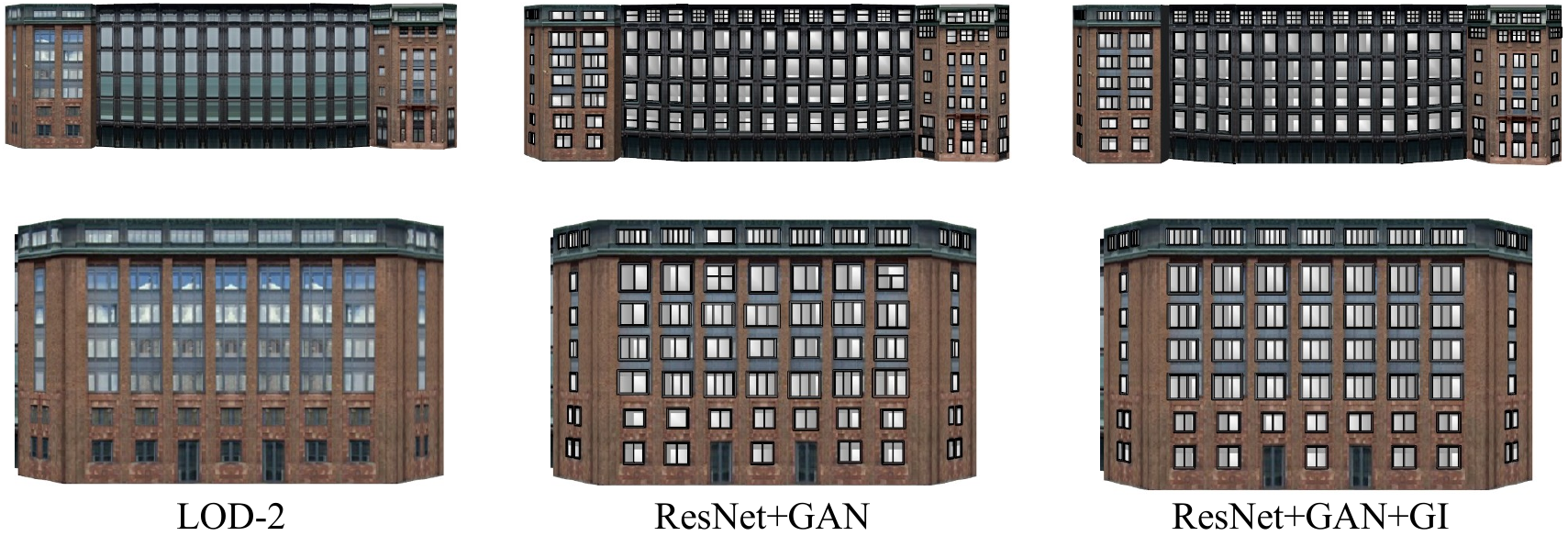}
	\caption{Reconstruction results for the Bracken House in the CBD dataset.}
	\label{fig:models_bracken}
\end{figure}

\begin{figure}[htb]
	\centering
	\includegraphics[width=0.8\linewidth]{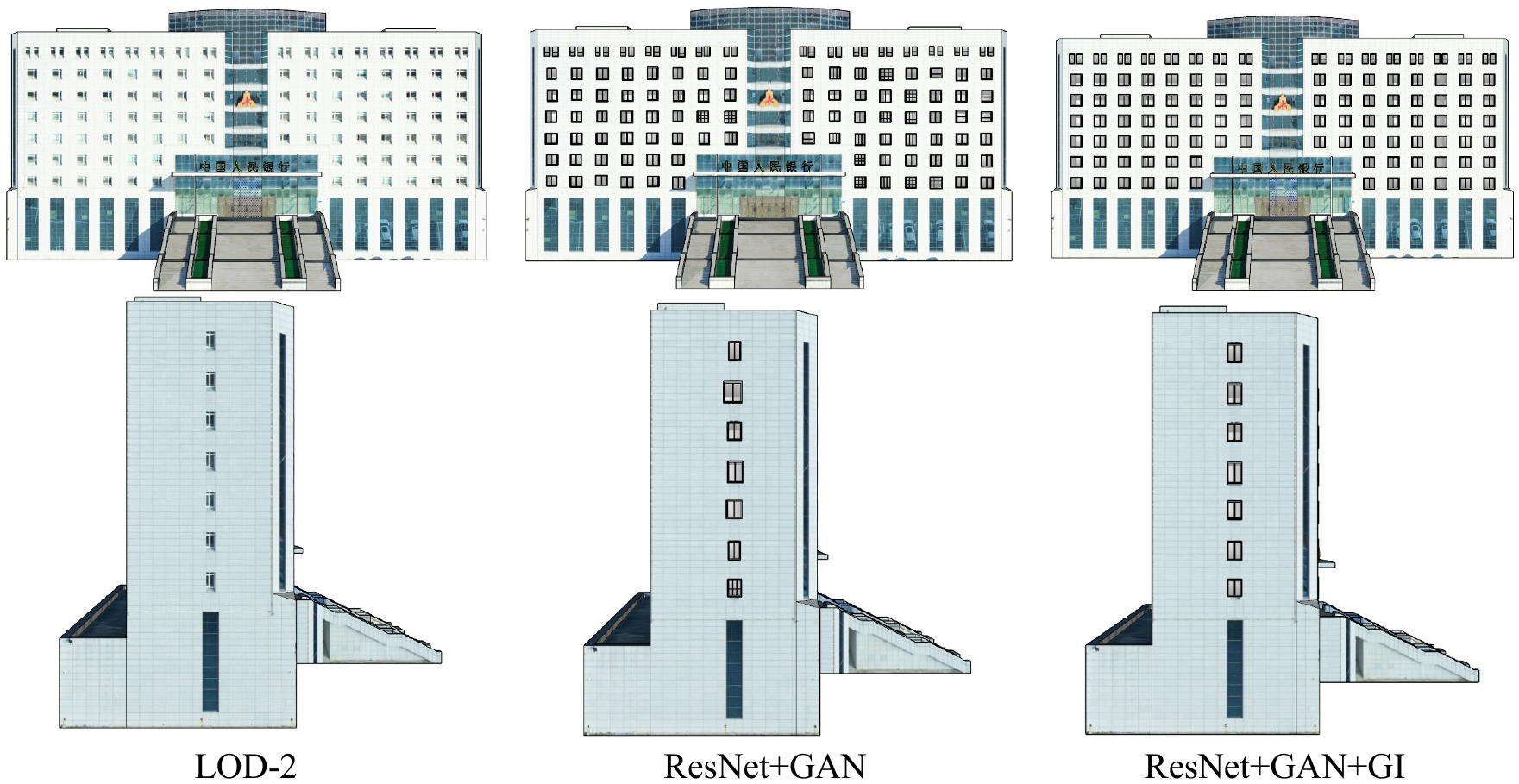}
	\caption{Reconstruction results for the Bank dataset.}
	\label{fig:models_bank}
\end{figure}

\begin{figure}[htb]
	\centering
	\includegraphics[width=0.8\linewidth]{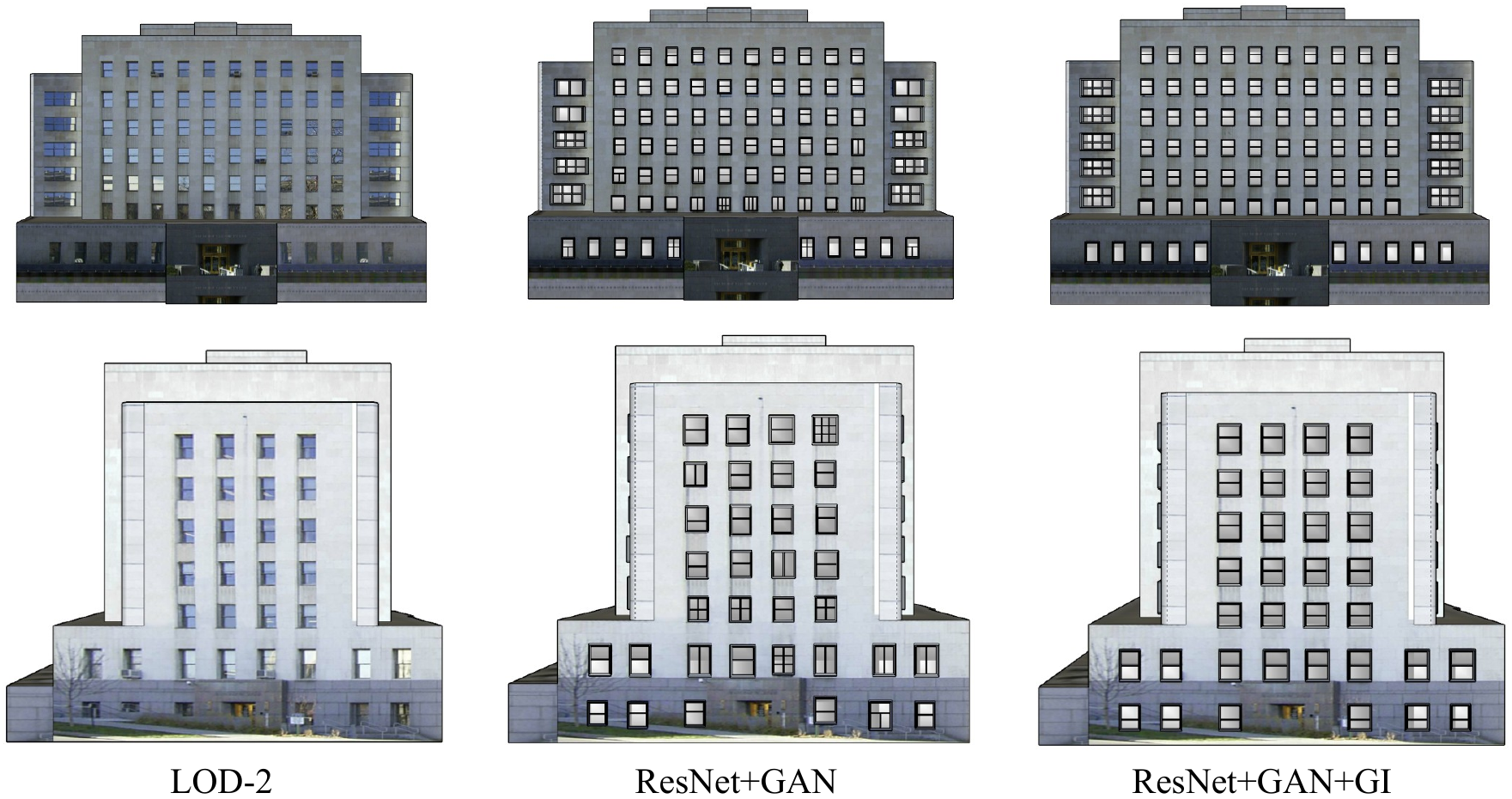}
	\caption{Reconstruction results for the State Capitol dataset.}
	\label{fig:models_capitol}
\end{figure}

To evaluate the proposed methods on various datasets, we also created several individual building models as shown in Figures \ref{fig:models_btcenter} through \ref{fig:models_capitol}.
Quantitative evaluations are displayed in Table \ref{tab:models}.
The only interactions are the grouping of windows for the same category.
For models with more repeated windows (BT Center and Bank), grouped inferences reach 100\% accuracy in classifying window types.
The models could be further improved by editing in details, assisted by the batching editing strategy, i.e., modifying only one instance and simultaneously synchronizing to all the instances.

Another noticeable characteristic is the generalization ability when trained with the adversarial strategy.
The bank dataset is significantly different from the training dataset, and vanilla ResNet performs only slightly better than random guess (1/9 in this case).
However, the proposed ResNet+GAN can still reach about 70\% Top-1 accuracy. 
More interactive representations and the downloadable models of results obtained with and without grouped inferencing are publicly available on the accompanying website.

\begin{table}[H]
	\centering
	\caption{Accuracies of the window types and parameters for four individual buildings.}
	\label{tab:models}
	\resizebox{\textwidth}{!}{%
		\begin{tabular}{@{}lccccc@{}}
			\toprule
			Dataset                                & Variant       & Top-1   (\%) & Top-2   (\%) & Top-3   (\%) & $\mathscr{L}_{MAE}$ \\ \midrule
			\multirow{3}{*}{BT   Center (CBD)}     & ResNet+GAN+GI & 100.00       & 100.00       & 100.00       & 17.60               \\
			& ResNet+GAN    & 88.30        & 98.83        & 99.42        & 17.62               \\
			& ResNet        & 54.39        & 78.37        & 92.40        & 38.11               \\ \midrule
			\multirow{3}{*}{Bracken   House (CBD)} & ResNet+GAN+GI & 82.39        & 93.18        & 93.18        & 27.21               \\
			& ResNet+GAN    & 62.50        & 92.61        & 94.32        & 28.81               \\
			& ResNet        & 57.39        & 75.00        & 83.52        & 50.95               \\ \midrule
			\multirow{3}{*}{Bank}                  & ResNet+GAN+GI & 100.00       & 100.00       & 100.00       & 31.32               \\
			& ResNet+GAN    & 67.91        & 86.10        & 93.05        & 31.91               \\
			& ResNet        & 17.65        & 27.27        & 31.02        & 40.94               \\ \midrule
			\multirow{3}{*}{State   Capitol}       & ResNet+GAN+GI & 82.46        & 82.46        & 91.23        & 19.64               \\
			& ResNet+GAN    & 71.05        & 84.21        & 92.98        & 20.04               \\
			& ResNet        & 78.95        & 85.09        & 87.72        & 40.94               \\ \bottomrule
		\end{tabular}%
	}
\end{table}

\subsection{Discussion and limitations}
Based on the evaluations above, we now return to the issues raised at the beginning of this paper.
In addition, we also discuss the limitations of the current strategy that could be surmounted in future works.

\textit{1) Breaking the domain gap in pre-training with few labeled data.
Fine-tuning a pre-trained network for domain-specific applications is a widely adopted strategy.
ImageNet or alternative large-scale vision datasets are generally utilized for pre-training.
However, if the domain-specific data, e.g., window images in our case, is not abundant, data-driven fine-tuning is not strong enough to pull the pre-trained weights from the source domain to the target domain.
To surmount the domain gap under a small number of labeled data, we adopt an additional pre-training step to transfer the weights from the ImageNet domain to the window domain with abundant unlabeled window patches.
Then, we conduct another fine-tuning step for the downstream tasks.}

\textit{2) Robust training from sparse samples.}
It is acknowledged that learned approaches are generally data-hungry.
Previous approaches either needed to construct a relatively large training database, i.e., exceeding $10^4$ samples \citep{nishida2018procedural}, or sought an appropriate strategy to simulate enough train samples \citep{zeng2018neural}.
We have shown that the GAN-based semi-supervised training can increase the robustness of the feature extractor.
We can achieve more than a 10\% recognition boost by replacing the pre-trained weights obtained from ImageNet with the one learned in this paper.
Because the testing time application can remain the same, the proposed adversarial training is a drop-in replacement in the standard pipeline.

\textit{3) Limitations.}
In order to decrease the complexity in the learning phase, we use only a single model to learn both the window types and grammar parameters.
The model could be fine-tuned into two separate branches, which is still significantly easier for integration than previous works \citep{nishida2016interactive}.
However, the flexibility for window grammar is also limited.
We only model windows with grid shapes.
Fortunately, the learned model can generalize gracefully to other window types.
For instance, the window with a curved half top will degrade to a window with two rows (Figure \ref{fig:model_borough}).

\section{Conclusion}
\label{s:conclusion}
Unlike the prosperity in object detection and semantic segmentation, learned approaches are less prevalent in the 3D modeling task.
A suitable connection between machine learning and 3D modeling is inverse procedural modeling or similar model-driven approaches.
The large number of training samples is the most important factor limiting applicability of deep learning methods.
This paper explored a semi-supervised training strategy to overcome the issue above.
We coined a prototype for detailed fa\c{c}ade reconstruction by embracing adversarial training and inverse procedural modeling.

Further research directions could combine the semi-supervised strategy with object detection \citep{ren2015faster}, which detects fine-grained window structures in an end-to-end manner.
In addition, more advanced generative networks \citep{zhu2020large} and semi-supervised techniques \citep{fan2022cossl,sohn2020fixmatch} are also interesting directions.

\section*{Acknowledgments}
This work was supported in part by the National Natural Science Foundation of China (Project No. 42071355, 41871291 and 41971411), in part by the National Key Research and Development Program of China (Project No. 2018YFC0825803), and in part by the Sichuan Science and Technology Program (Project No. 2020JDTD0003).

\bibliographystyle{model2-names}
\bibliography{FacadeWindow}

\end{document}